\documentclass{article} 
\usepackage{iclr2026_conference,times}

\usepackage{amsmath,amsfonts,bm}
\usepackage{graphicx}
\usepackage{mathtools}

\def\eqref#1{equation~\ref{#1}}

\def\1{\bm{1}}

\DeclareMathAlphabet{\mathsfit}{\encodingdefault}{\sfdefault}{m}{sl}
\SetMathAlphabet{\mathsfit}{bold}{\encodingdefault}{\sfdefault}{bx}{n}

\usepackage{algorithm}
\usepackage[noend]{algpseudocode}
\usepackage{hyperref}
\usepackage{booktabs}    
\usepackage{url}
\usepackage{colortbl}
\usepackage{xcolor, etoolbox}
\usepackage{float}
\usepackage[inkscapelatex=false]{svg}
\usepackage{wrapfig}
\usepackage{makecell} 
\usepackage{subcaption}
\usepackage{graphicx}
\usepackage[table]{xcolor}
\usepackage{tikz}
\usepackage{amssymb}
\usepackage{todonotes}
\usepackage{multirow}
\usepackage{orcidlink}
\usetikzlibrary{fit,calc}
\definecolor{forestgreen}{RGB}{34,139,34}
\definecolor{olivegreen}{RGB}{107,142,35}
\definecolor{limegreen}{RGB}{50,205,50}
\usepackage{pifont} 
\newcommand{\cmark}{\textcolor{forestgreen}{\ding{51}}}
\newcommand{\xmark}{\textcolor{red}{\ding{55}}}

\usepackage[table]{xcolor}
\definecolor{oursrowcolor}{RGB}{230,245,255}
\newcommand{\oursrow}{\rowcolor{oursrowcolor}}

\colorlet{mypink}{red!40}
\colorlet{myblue}{cyan!60}
\title{CRONOS:  Continuous time reconstruction for 4D medical longitudinal series } 

\newcommand{\TFM}{TFM}
\newcommand{\ctfm}{Continuous RecOnstructioNs for medical lOngitudinal Series}
\newcommand{\CTFM}{CRONOS}

\newcommand{\ib}{Last Context Image}
\newcommand{\IB}{LCI}

\newcommand{\timt}{t_\text{target}}
\newcommand{\Imgt}{\mathcal{I}_\text{target}}
\author{Nico Albert Disch$^{1,2,3}$\orcidlink{0000-0001-8791-622x} \And Saikat Roy$^{1,3}$\orcidlink{0000-0002-0809-6524} \And Constantin Ulrich$^{1,5}$\orcidlink{0000-0003-3002-8170} \And Yannick Kirchhoff$^{1,2,3}$\orcidlink{0000-0001-8124-8435} \And Maximilian Rokuss$^{1,3}$\orcidlink{0009-0004-4560-0760} \And Robin Peretzke$^{1,5}$\orcidlink{0000-0002-6187-3636} \And David Zimmerer$^{1,2}$\orcidlink{0000-0002-8865-2171} \AND Klaus Maier-Hein$^{1,2,4,6}$\orcidlink{0000-0002-6626-2463} \\ 
$^{1}$ Division of Medical Image Computing, German Cancer Research Center, Heidelberg, Germany  \\
$^{2}$ HIDSS4Health - Helmholtz Information and Data Science School for Health,\\Karlsruhe/Heidelberg, Germany \\
$^{3}$ Faculty of Mathematics and Computer Science, University of Heidelberg, Heidelberg, Germany \\
$^{4}$ Pattern Analysis and Learning Group, Department of Radiation Oncology, 
\\ Heidelberg University Hospital, Heidelberg, Germany \\
$^{5}$ Medical Faculty Heidelberg, University of Heidelberg, Heidelberg, Germany \\
$^{6}$ Pattern Analysis and Learning Group, Department of Radiation Oncology\\ Heidelberg University Hospital \\
\texttt{\{nico.disch\}@dkfz.de} \\
}
\iclrfinalcopy 
\begin{document}

\maketitle

\begin{abstract}
Forecasting how 3D medical scans evolve over time is important for disease progression, treatment planning, and developmental assessment.
Yet existing models either rely on a single prior scan, fixed grid times, or target global labels, which limits voxel-level forecasting under irregular sampling.
We present \CTFM{}, a unified framework for many-to-one prediction from multiple past scans that supports both discrete (grid-based) and continuous (real-valued) timestamps in one model, to the best of our knowledge the first to achieve continuous sequence-to-image forecasting for 3D medical data.
\CTFM{} learns a spatio-temporal velocity field that transports context volumes toward a target volume at an arbitrary time, while operating directly in 3D voxel space.
Across three public datasets spanning Cine-MRI, perfusion CT, and longitudinal MRI, \CTFM{} outperforms other baselines, while remaining computationally competitive. 
We will release code and evaluation protocols to enable reproducible, multi-dataset benchmarking of multi-context, continuous-time forecasting. \thanks{ Code will be made available under \url{https://github.com/MIC-DKFZ/Longitudinal4DMed}}
\end{abstract}
\section{Introduction}
\captionsetup[subfigure]{labelfont=bf} 
\begin{figure}[!htpb]
  \centering
  \begin{subfigure}{0.48\linewidth}
    \centering
    \includegraphics[width=\linewidth]{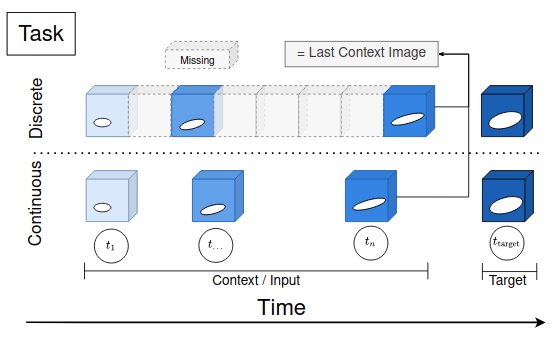}
    \caption{Task description.}
    \label{fig:schematic}
  \end{subfigure}\hfill
  \begin{subfigure}{0.48\linewidth}
    \centering
    \includegraphics[width=\linewidth]{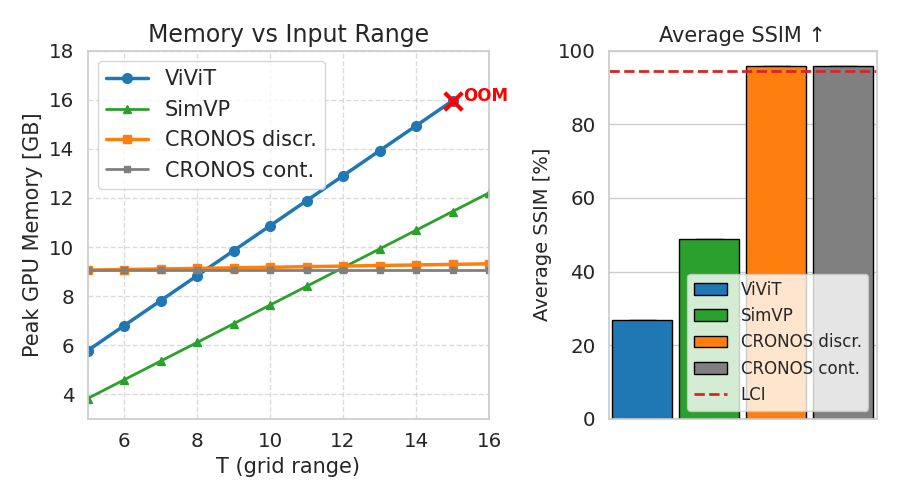}
    \caption{Memory Efficiency vs. Performance}
    \label{fig:efficiency}
  \end{subfigure}
\caption{
\textbf{Task and benchmark comparison}
\textbf{(a) Task setup} Forecasting a target 3D scan from multiple past volumes in two regimes. 
\textit{Discrete:} acquisitions lie approximately on a regular grid, but may contain missing frames (dotted boxes).
\textit{Continuous:} acquisitions occur at irregular, real-valued timestamps and are used directly without grid alignment.
Many-to-one task $(\{I_i\}_{i=1}^T, t_\text{target}) \to I_\text{target}$.
\textbf{(b) Efficiency and performance} Left: GPU memory scaling of single forward pass with sequence length $T$ shows \CTFM{} to be substantially more memory-efficient than alternatives. Right: Average SSIM across two datasets, where \CTFM{} outperforms baselines and \IB{}.
}
  \label{fig:task_and_model}
\end{figure}
Longitudinal medical imaging is central to monitoring disease progression, assessing treatment response, and modeling anatomical development across time \citep{suter_lumiere_2022, rivail_modeling_2019, bernard_deep_2018}.
Some modalities  are inherently spatio-temporal, such as ultrasound (US), cine-MRI, videos, or perfusion Computer Tomography (CT).
Beyond these, repeated clinical acquisitions form temporal sequences that may span over months or years and are used for clinical decision making.
In ophthalmology, for instance, longitudinal OCT volumes are central to monitoring progression of age-related macular degeneration and predicting treatment response~\citep{rivail_modeling_2019}.
Works such as using surgical video streams ~\citep{li_deep_2024}, which are also increasingly leveraged for diverse tasks, or in ~\citep{gomes_ai_2022}, where longitudinal US sequences are used, show the overall breadth of spatio-temporal imaging.
\begin{wraptable}{r}{0.6\textwidth} 
\centering
\begin{tabular}{llcccc}
\toprule 
\textbf{Category} & \textbf{Method} & \textbf{C1} & \textbf{C2} & \textbf{C3} & \textbf{C4} \\
\midrule
\multirow{3}{*}{Med. Gen} 
    & BrLP & \xmark & \cmark & \cmark & \cmark \\  
    & LociDiffCom & \xmark & \cmark & \cmark & \xmark \\  
    & ImageFlowNet & \xmark & \cmark & \xmark & \cmark \\  
\midrule
\multirow{1}{*}{Vid. Gen} 
    & MCVD & \cmark & \cmark & \xmark & \xmark \\  
\midrule
\multirow{4}{*}{STL}
  & SimVP & \cmark & \xmark & \cmark & \xmark \\
  & ViViT & \cmark & \xmark & \cmark & \xmark \\
  & ConvLSTM & \cmark & \xmark & \cmark & \xmark \\
  &NODE+LSTM & \cmark & \xmark & \cmark & \cmark \\
\midrule
\multirow{1}{*}{Med. STL}
  & \CTFM{} (ours) & \cmark & \cmark & \cmark & \cmark \\
\bottomrule
\end{tabular}
\caption{\textbf{Technical comparison of spatio-temporal prediction methods.} 
Columns denote Challenges ($C \#$): \textbf{C1: Multiple Inputs}, \textbf{C2: high fidelity}, \textbf{C3: 3D imaging}, \textbf{C4: continuous-time modeling}. 
Our proposed \CTFM{} satisfies all four criteria, whereas existing medical and natural imaging baselines lack one or more.
STL stands for spatio-temporal learning.
}
\label{tab:comparison_other_works}
\vspace{-0.5cm}
\end{wraptable}
Beyond individual modalities, there is also a massive and growing amount of video and longitudinal data across clinical contexts~\citep{farhad_review_2023}, including applications such as treatment response prediction in oncology ~\citep{suter_lumiere_2022}.

Despite its importance, spatio-temporal learning in medical imaging is centered mostly on single time-point (image-to-image) analysis. 
Some approaches rely on global labels e.g.~\cite{yoon_latent_2024}, while many reduce to image-to-image preidction with a single context scan \citep{zhang_diffusion_2025}).
 Ohters introduce task-specific prior or remain tied to one disease (e.g.~\cite{puglisi_brain_2025}.
In particular, Alzheimer's Disease (AD) has attracted a disproportionate share of longitudinal imaging research~\citep{petersen_alzheimers_2010,marti-juan_survey_2020,chen_reflections_2025}, whereas other domains remain comparatively underexplored.

\CTFM{} addresses these challenges by introducing a unified spatio-temporal flow framework for medical sequence-to-image prediction that: \footnote{Code will be released at \url{github.com/anonymous}.}
\begin{itemize}
    \item \textbf{Supports both \emph{discrete} and \emph{continuous} timestamps}, leveraging multiple past scans jointly on \textbf{3D} medical imaging data.
    \item \textbf{Avoids disease-specific assumptions}, enabling application to any medical longitudinal task.
    \item \textbf{Consistently outperforms prior approaches}, including standard sequence models and the  \ib{} (\IB{}) baseline, which is a surprisingly simple and competitive heuristic (NRMSE, PSNR, and SSIM), due to slowly changing medical images.
\end{itemize}

\section{Related work}
\paragraph{Medical Imaging}
Prior work in longitudinal medical imaging focuses heavily on one-to-one, or one-to-many video prediction.
While approaches like diffusion models~\citep{linguraru_tadm_2024,zhu_loci-diffcom_2024,puglisi_brain_2025} and Neural ODEs~\citep{lachinov_learning_2022,liu_imageflownet_2025} have been applied to medical imaging,
these are \emph{image-to-image}, and thus cannot canonically capture multi-input longitudinal evolution. 
For example, \cite{bai_noder_2024} propose a continuous-time model, but they predict sequences from single images. 
In contrast, works that jointly leverage multiple observations show improved prediction accuracy~\citep{fang_deep_2021}. 
The single-context nature makes these aforementioned works not sufficient for our setting.
There are also \textbf{interpolation-based methods} ~\citep{linguraru_loci-diffcom_2024} which predict intermediate frames between two acquisitions, but this restricts their use to filling missing intervals rather than forecasting. 
Overall, existing medical approaches are all technically restricted; be it only single-image input, disease specific priors, limited to 2D, or not being able to forecast to arbitrary times as shown in ~\ref{tab:comparison_other_works}. 
\begin{figure}[H]
    \centering
    \includegraphics[width=0.85\linewidth]{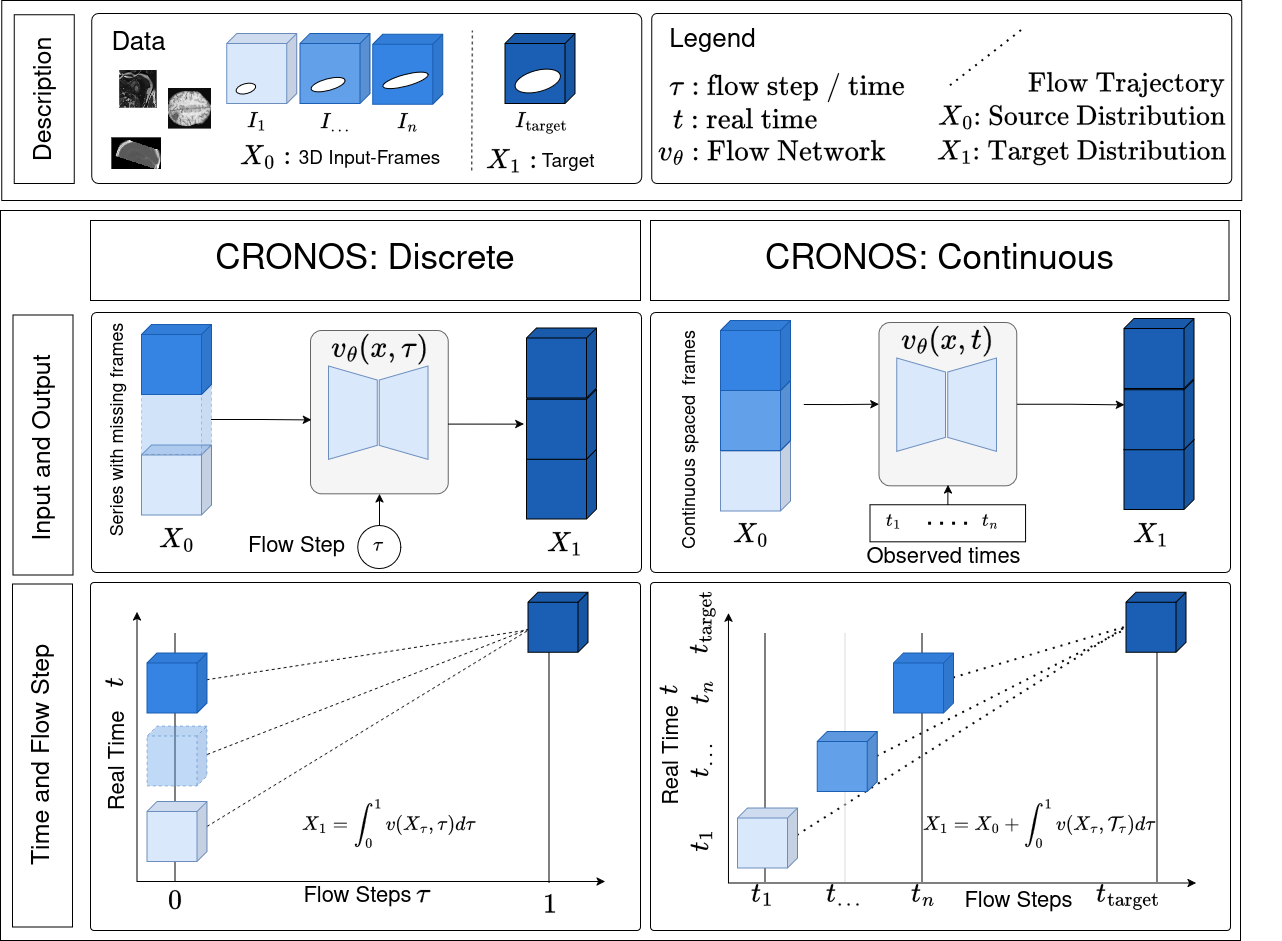}
        \caption{\textbf{\CTFM{} method overview:}
\textbf{Left:} Discrete \CTFM{} treats time implicitly, interpolating between context frames and a fixed target along a normalized flow step $t \in [0,1]$. 
\textbf{Right:} Continuous \CTFM{} explicitly conditions on real-valued timestamps $t_i$, allowing each context $I_i$ to transport toward the target via its own interpolation $t_i$. This enables predictions at arbitrary target times while preserving the true temporal geometry.}
\label{fig:time_embedding_figure}
\end{figure}
By contrast, our work focuses on continuous-time modeling across full spatio-temporal sequences without restricting to specific modalities or diseases.
\paragraph{Natural Imaging and Video Prediction}
Spatio-temporal modeling has been extensively studied in video prediction. Early approaches such as ConvLSTM~\citep{shi_convolutional_2015} introduced recurrent sequence-to-sequence architectures and remain widely used. Subsequent methods such as SimVP~\citep{gao_simvp_2022} replaced recurrence with purely convolutional designs. Transformer-based models like ViViT~\citep{arnab_vivit_2021} extended attention mechanisms to the video domain and have become a backbone in many imaging domains. More recent efforts have explored generative modeling, including video diffusion~\citep{voleti_mcvd_2022,ye_stdiff_2023,yan_videogpt_2021}, and continuous-time formulations such as Neural ODEs~\citep{chen_neural_2019}, extended to videos in~\citep{park_vid-ode_2021}. 
While these approaches are powerful, they have primarily been developed for dense 2D natural video sequences with large-scale training data.
Accordingly, they transfer poorly to 3D medical images with small datasets and sparse sequences , thus motivating our work. 
\paragraph{Flow Matching}
Flow Matching (FM) has recently emerged as a generative modeling paradigm~\citep{lipman_flow_2023,lipman_flow_2024}, and has been adapted to irregular time series, e.g. in~\citep{zhang_trajectory_2025}, though only for low-dimensional data rather than full image sequences. Our extension  therefore is: while classical FM learn a single flow from (most often) raw noise $X_0\!\sim\! p $ to samples $X_1\!\sim\! 1$ along steps $\tau \in [0,1]$, we re-cast
\begin{equation}
    X_0=[I_1,\dots,I_T],\qquad X_1=\Imgt:=[I_{\text{target}}, \dots, I_{\text{target}}] ,
\end{equation}
interpreting $p$ as the context sequence, and $q$ as a broadcast stack of $I_\text{target}$ ( defined the stack as $\Imgt$, to make dimension explicit).
This temporal broadcasting turns FM into sequence-to-image transport: a shared velocity field $v_\theta$ simultaneously moves all $T$ context volumes toward the target, effectively $T$ per-frame transports under shared parameters.
We refer to this framework as \textbf{\ctfm{} (\CTFM{})}.

\section{Methods}\label{sec:methods}
\begin{algorithm}[!htb]
  \caption{\CTFM{} Continuous: Training and Inference}
  \label{alg:continuous_tfm}
  \begin{algorithmic}[1]
    \Require Patients $\mathcal{P}$ and initial network $v_\theta$
    \While{training}
      \State Sample $\{[\mathcal{I}, I_{\mathrm{target}}],\,[t_1, \dots, t_T, \timt]\} \sim \mathcal{P}(\mathcal{X})$ \Comment{pick a random patient}
      \State Sample $\tau \sim \mathcal{U}(0,1)$ \Comment{random flow step}
      \State $\mathcal{I}_\mathrm{target}\gets [I_\mathrm{target},\dots, I_\mathrm{target}]$ \Comment{repeat target $T$ times}
      \State $\mathcal{T}'_\tau\gets (1-\tau)\,[t_1, \dots, t_n] + \tau\,\bm{t}_{\mathrm{target}}$ \Comment{interpolate timestamps}
      \State $X_\tau\gets (1-\tau)\,\mathcal{I} + \tau\,\mathcal{I}_\mathrm{target}  + \sigma (\tau) \epsilon $\Comment{linear interpolation}
      \State $\mathcal{L}\gets \|v_\theta(\mathcal{T}'_\tau,X_\tau) - (\mathcal{I}_\mathrm{target}-\mathcal{I})\|^2$ \Comment{velocity loss}
      \State Update $\theta\leftarrow\mathrm{AdamW}(\nabla_\theta\mathcal{L})$
    \EndWhile
    \State \Return $v_\theta$
    \If{inference}
      \State Initialize $X_0\gets\mathcal{I}$
      \State Define integration grid $\{\tau_0=0,\dots,\tau_N=1\}$ with $N$ steps
      \State $\mathcal{T}'_\tau = (1-\tau)\,[t_1, \dots, t_n] + \tau\,\bm{t}_{\mathrm{target}}$
      \State $\hat{X}_{0:N}\gets \mathrm{ODEInt}(v_\theta,X_0,\{\mathcal{T}'_0,\dots,\mathcal{T}'_1\})$ \Comment{numerical integration}
      \State \Return $\hat{X}_N$
    \EndIf
  \end{algorithmic}
\end{algorithm}

\subsection{Problem Setup}\label{subsec:problem_setup}

\begin{wrapfigure}{l}{0.50\textwidth}
    \centering
    \includegraphics[width=1\linewidth]{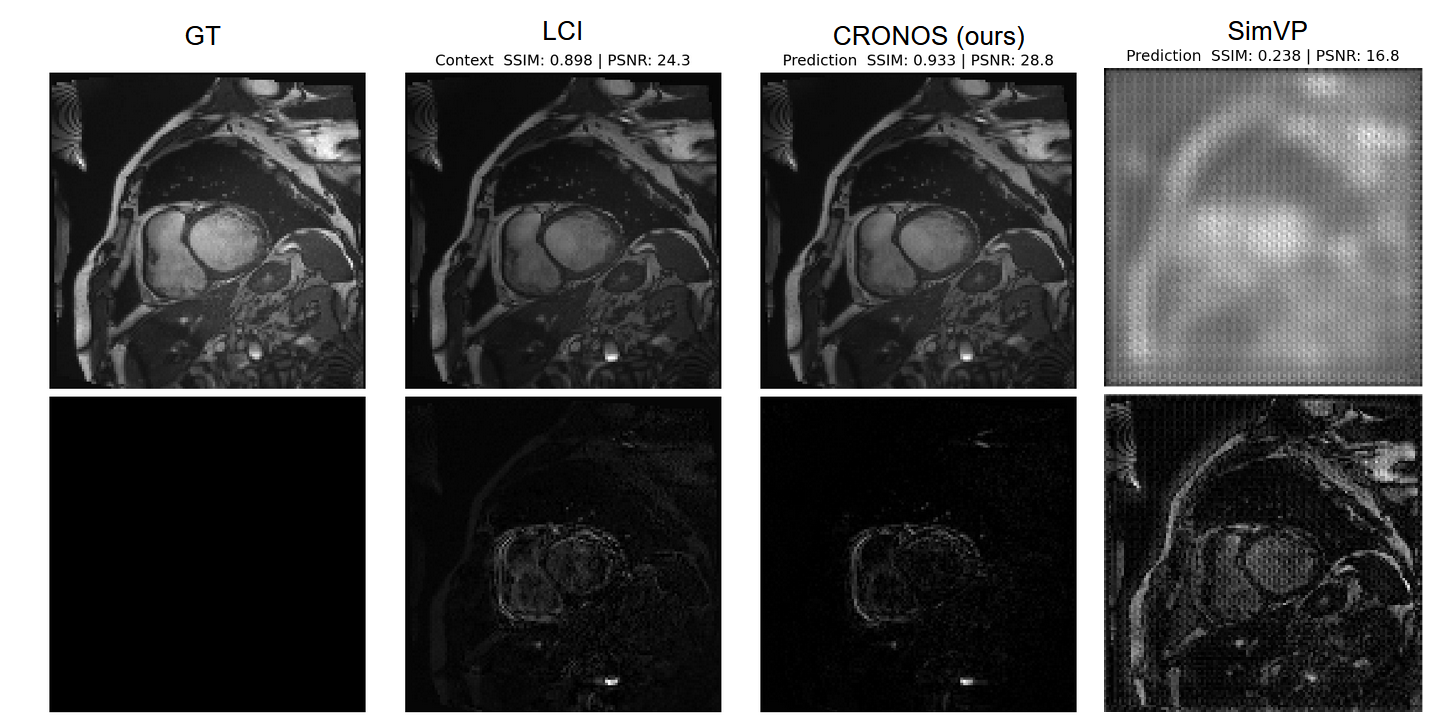}
    \caption{\textbf{Qualitative comparison on the ACDC dataset}. Ground truth (GT), Last Context Image (LCI), our method (\CTFM{}), and SimVP. 
    Upper row: prediction, lower row: residuals.
}
    \label{fig:qualitative_acdc}
\end{wrapfigure}
Let $\mathcal{P} = \big\{\,(\{I_i^{(n)}, t_i^{(n)}\}_{i=1}^{T^{(n)}},\; t_{\mathrm{target}}^{(n)},\; I_{\mathrm{target}}^{(n)})\,\big\}_{n=1}^p$ denote a dataset of $p$ patient sequences.
Each (patient) sequence consists of a set of $T$ context volumes $\mathcal{I} = \{I_1,\dots,I_T\}$, with $I_i\in\mathbb{R}^{H\times D\times W}$ (for shorthand $S=H\times D\times W$),
acquired at associated timestamps $\{t_1,\dots,t_T\}\subset\mathbb{R}_+$.
We consider two regimes. 
\textit{\textbf{Discrete}}: Acquisitions lie on a uniform time grid; some frames may be missing, yielding sparse sequences (e.g., natural video, cine-MRI, perfusion CT). 
\textit{\textbf{Continuous}}: Acquisitions occur at irregular, real-valued times that do not easily align to any grid (typical in longitudinal clinical scans). 
For continuous series, forcing a frame grid either explodes sequence length with empty slots or loses temporal precision.
For instance, daily-resolution for timepoints over several years would yield $T$ in the thousands, yet in practice only a handful of scans are ever acquired. 
In both discrete and continuous series, $T$ is small relative to natural video.

\emph{Target Task.} Given the set of context images and time$\{(I_i,t_i)\}_{i=1}^T$, as well as a target time $t_{\mathrm{target}}$, we aim to learn
\begin{equation}\label{eq:mapping_function}
    f\big(\{I_i,t_i\}_{i=1}^T,\; t_{\mathrm{target}}\big) \mapsto I_{\mathrm{target}}.
\end{equation}
The discrete setting uses a fixed grid (with optional zero-tensors for missing context volumes);
the continuous setting uses the \textbf{observed} context only, without padding.

\subsection{Flow Matching (FM)}\label{subsec:fm}
Flow Matching~\cite{lipman_flow_2023} learns a ordinary differential equation (ODE), linking the equal dimensional distributions $p$ and $q$ via 
\begin{equation}\label{eq:fm_ode}
    \frac{d}{d\tau}\psi_\tau(x) = u_\tau(\psi_\tau(x)),\qquad X_1 = X_0 + \int_0^1 u_\tau(X_\tau)\,d\tau,
\end{equation}
with $X_0\!\sim\!p$, $X_1\!\sim\!q$. 
A convenient coupling is obtained by sampling $X_\tau$ as 
\begin{equation}\label{eq:fm_path}
    X_\tau = (1-\tau) X_0 + \tau X_1  + \sigma (\tau) \epsilon,
\end{equation}
where  $\epsilon \sim \mathcal{N}(0,I)$ denotes random gaussian noise and $\sigma (\tau)$ its intensity, which is sampled around the straight path.
The corresponding ground-truth velocity along this path is therefore constant:
\begin{equation}\label{eq:fm_u}
    u_\tau(X_\tau) = \frac{d}{d\tau}X_\tau = X_1 - X_0.
\end{equation}
Consequently, to approximate the ground truth velocity, we train a neural network $v_\theta(X_\tau,\tau) \in \mathbb{R}^{T \times S}$ using:
\begin{equation}\label{eq:fm_loss}
    \mathcal{L}_{\mathrm{CFM}}=\mathbb{E}_{X_0,X_1,\tau}\big\|v_\theta(X_\tau,\tau)-u_\tau(X_\tau)\big\|_2^2.
\end{equation}
Using $v_\theta$, we can then infer using ~\eqref{eq:fm_ode} via an approximate ODE solver.

\subsection{Continuous and discrete reconstructions for medical image time series (\CTFM{})}\label{subsec:ctfm}
We introduce \CTFM{}, a spatio-temporal flow model that learns continuous trajectories from longitudinal scans. 
It comes in two complementary variants: \emph{discrete} and \emph{continuous}.

\paragraph{Temporal broadcasting for sequence-to-image flows}
To enable flow between a sequence of context images and a single target, we define $X_0 \sim p$ as the stack of context images (with variant-specific handling for continuous vs. discrete), and $X_1\sim q$ as the target image broadcast to the same shape
\begin{equation}\label{eq:broadcast}
    X_1 = [I_{\mathrm{target}},\dots,I_{\mathrm{target}}].
\end{equation}
This broadcasting ensures that $X_0$ and $X_1$ share the same dimensionality, allowing us to define a valid flow between them.

\paragraph{Discrete \CTFM{}.}  
On a regular grid with missing scans, we first \textit{embed} each sequence onto the grid of a resolution $\mathrm{g}$ using a binning operator  $\mathcal{E}^{\mathrm{grid}}_{\mathbf g}$, which assigns each $I_i$ to the closes grid index matching $t_i$ (proper definition in ~\ref{app:grid_embedding}).
Missing slots are then handled by a last-observed carry-forward operator  $\mathcal{F}^{\mathrm{LOCF}}$, which fills empty positions with the most recent available scan.
In short, we define
\begin{equation}
    X_{0}
= \big(\underbrace{\mathcal{F}^{\mathrm{LOCF}}}_{\text{fill}}
  \circ \underbrace{\mathcal{E}^{\mathrm{grid}}_{\mathbf g}}_{\text{bin to grid}}\big)
  \!\left(\{(I_i,t_i)\}_{i=1}^T\right)
= [\hat I_1,\dots,\hat I_K].
\end{equation}
This pre-processing ensures $X_0$ is well-defined on a uniform grid.
Furthermore, $\mathrm{LOCF}$ handles spatial missingness: missing frames are zero-initialized and replaced by the most recent observation (Appendix ~\ref{app:sparsity_filling}).
This setup stabilizes optimization and preserves grid order while enabling many-to-one sequence transport within FM.
Finally, we train on the linear interpolation
$X_\tau=(1-\tau)X_0+\tau X_1$ using \eqref{eq:fm_loss}, where temporal order is captured \emph{implicitly} by the flow step $\tau$ and the frame index.
Additionally, we set $\sigma = 0$ during training and inference, ablations on nonzero noise levels are reported in Table~\ref{tab:ablation_training_noise}.
\paragraph{Continuous \CTFM{}}
Our continuous modeling strategy extends on the discrete case by conditioning on \emph{real-valued timestamps} while evolving along a scalar flow parameter $\tau\in[0,1]$.
We construct spatio-temporal tensors (using mild abuse of notation):
Time enters the network only as conditioning on real timestamps.
We interpolate the conditioning timestamps along the interpolated time vector $\mathcal{T}_\tau$.
$X_0$ is then defined as in equation ~\eqref{eq:broadcast}, without embedding it to the grid, nor performing LOCF as in discrete \CTFM{}. 
We define the shifted time vector as
\begin{equation}\label{eq:tinterp}
    \mathcal{T}_\tau = (1-\tau)\,\mathbf{t}_{\mathrm{ctx}} + \tau\,\mathbf{t}_{\mathrm{target}}.
\end{equation}
The formulation in ~\eqref{eq:tinterp} lets flow step $\tau$ carry real temporal information, without adding extra complexity.
The conditional trajectory is then
\begin{equation}\label{eq:ctfm_ode}
    X_{1} \;=\; X_{0} + \int_{0}^{1} v_{\theta}\!\big(X_{\tau},\, \mathcal{T}_{\tau}\big)\, d\tau ,
\end{equation}
where $v_{\theta}$ is the predicted velocity field and \emph{$\tau$ is the flow step} (usually called time, we avoid it due to avoiding confusion).
Prediction is then done via approximate solution of ~\eqref{eq:fm_ode}, solver details found in ~\ref{subsec:exp_details}.
This formulation lets \CTFM{} model continuous image evolution grounded in actual scan times, supporting interpolation or forecasting without regular sampling or artificial frame filling.
It avoids zero-padding, leading to reduced computational burden compared to the discrete variant.
Both variants use the same 3D U-Net backbone, further details are provided in Appendix~\ref{subsec:network_architecture}, and the training/inference procedure appears in Algorithm~\ref{alg:continuous_tfm}.
\paragraph{Time Encoding.}
Flow steps and continuous times are mapped to Fourier embeddings using  ~\citep{tancik_fourier_2020}, which were used e.g. in \citep{rombach_high-resolution_2022}:
$\gamma(t) = [\sin(2\pi f_k t), \cos(2\pi f_k t)]_{k=1}^K$ using frequencies $f_k$.  
To preserve dimensional consistency across variable-length input sequences for the continuous setting, we  compute the time embedding as  
\begin{equation}
    \text{Enc}(\bm{t}) = \frac{1}{T}\sum_{i=1}^T \gamma(t_i).
\end{equation}
This embedding is then added to each residual layer via FiLM. 
The loss is then calculated via ~\eqref{eq:fm_loss}, and inference via ~\eqref{eq:ctfm_ode}.

\section{Data and Experimental Design}\label{sec:data-and-experimental-design}
\subsection{Datasets}\label{subsec:datasets}
\begin{wrapfigure}{r}{0.35\linewidth}
    \centering
    \includegraphics[width=1\linewidth]{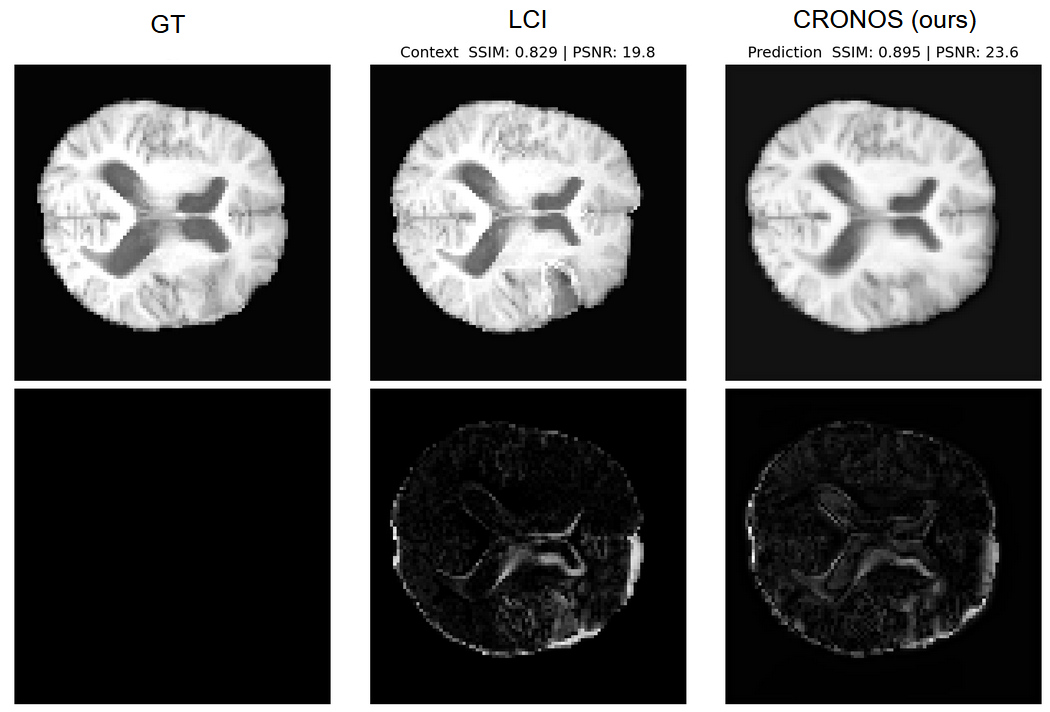}
    \caption{\textbf{Qualitative comparison on the LUMIERE dataset}. Ground truth (GT), Last Context Image (LCI), our method (\CTFM{}), and SimVP baseline. 
Lumiere is particularly challenging due the very small dataset. 
highlighting the benefit of explicit continuous-time conditioning under extreme data scarcity.}
\label{fig:lumiere_qualitative}
\end{wrapfigure}

\textbf{ACDC}~\citep{bernard_deep_2018} is a cardiac MRI dataset capturing different heart phases.
The context tensor is reshaped to $[T,H,D,W]=[11,32,128,128]$, and the target is a single image with the same spatial size.
We split ACDC into $80$ training, $20$ validation and $50$ test images.
This dataset served for method development; ablations were conducted on the validation split.
\\ \noindent
\textbf{ISLES} ~\citep{riedel_isles_2024} consists of perfusion CT image time series from stroke patients.
From the normalized series, we sample $7$ consecutive points, take the last as the target, and randomly mask the remaining context frames.
The resulting context tensor has shape $[T,H,D,W]=[7,16,128,128]$.
We use a split of $92$ training, $23$ validation and $34$ test images.
For both the ACDC and dataset, we randomly mask out time points (see Appendix ~\ref{subsec:random_masking}).
\\ \noindent
\textbf{Lumiere}~\citep{suter_lumiere_2022} is a longitudinal glioma MRI dataset with 3D scans.
Images are reshaped to $[T,H,D,W]=[7,96,96,64]$.
Because some patients have few acquisitions, we prepend zeros to standardize pre-processing across cases.
The split is $48$ training, $12$ validation and $14$ test images.

\subsection{Experimental settings}\label{subsec:experimental-settings}
Reproducibility details can be found in Section ~\ref{sec:reproducibility}.

\textbf{Discrete Setting}: As mentioned in the data section, input data has dimension $T$, while some frames may be missing. 
We apply \textit{both} variants of \CTFM{}, noting that the continuous version can also operate in this regime with a smaller context window, since missing images \textit{do not need} to be explicitly represented. 
The lower context window also leads to a lower computational demand.
Therefore, the underlying tensors remain uniform, with some time points masked. 
For validation and testing we ensure that the missingness pattern is fixed across epochs, as otherwise the choice of best checkpoint would be ill-posed (further details in Appendix~\ref{subsec:random_masking}).

\textbf{Continuous Setting}:
As an \textit{additional ablation and experiment}, we simulate a continuous setup on ACDC to highlight the gains from explicit timestamp conditioning.
While no public dataset provides plenty of continuous acquisition protocols, this sub-sampled variant shows that \CTFM{} benefits from real-valued time even beyond irregular masking.
Specific details of how we subsampled ACDC can be found in ~\ref{subsec:exp_details}.
Importantly, both the discrete and continuous formulations remain applicable to discrete grids.

\begin{table}[!h]
\centering
\caption{\textbf{Discrete Time: Quantitative Evaluation on Many-to-One Sequences:} Reported values are mean (standard deviation) over three runs. 
Metrics include normalized root $MSE$, $NRMSE$, structural similarity index ($SSIM [\%]$) and peak signal-to-noise-ratio $PSNR$.
*ViViT OOM on a 40 GB GPU, despite having a smaller batch size and the lowest possible feature size. Standard deviation of \IB{} omitted for visual clarity. Blue row: only method to beat \IB{} and our proposed \CTFM{}. Computational requirements on ACDC in ~\ref{sec:appendix}}
\label{tab:main_results}
\begin{tabular}{l l l l l}
\toprule
Dataset & Model &  NRMSE [$10^{-2}$] $\downarrow$ & SSIM [$\%$] $\uparrow$ & PSNR [$dB$] $\uparrow$  \\
\midrule
\multirow[t]{7}{*}{ACDC}& LCI & \phantom{0}4.48 & 92.79 & 28.918 \\
\cline{2-5}
& ConvLSTM & 11.20 ± 0.48 & 50.44 ± 1.53 & 19.123 ± 0.312 \\ 
 & SimVP & \phantom{0}9.27 ± 0.29 & 49.08 ± 4.01 & 20.715 ± 0.267 \\
 & NODE + LSTM & 11.59 ± 0.18 & 36.41 ± 2.94 & 18.946 ± 0.186 \\
 & ViViT & 13.90 ± 2.66 & 17.06 ± 8.60 & 17.252 ± 1.738 \\
 \oursrow
 &\CTFM{} discrete & \phantom{0}\underline{3.97 ± 1.23} & \textbf{94.51 ± 0.79} & \textbf{30.510 ± 1.560} \\
 \oursrow
 & \CTFM{} cont. & \phantom{0}\textbf{3.74 ± 0.21} & \underline{94.34 ± 0.45} & \underline{29.750 ± 0.528} \\
\midrule
\multirow[t]{7}{*}{ISLES} & LCI & \phantom{0}5.25 & 96.29 & 29.002 \\
\cline{2-5}
& ConvLSTM & 19.31 ± 0.18 & 39.92 ± 0.66 & 17.644 ± 0.014 \\ 
 & SimVP & 13.06 ± 0.19 & 48.82 ± 1.60 & 20.799 ± 0.112 \\
 & ViViT & 16.54 ± 0.30 & 36.76 ± 1.49 & 18.671 ± 0.134 \\
 & NODE + LSTM & 15.10 ± 0.87 & 40.55 ± 7.15 & 19.481 ± 0.515 \\
  \oursrow
 &\CTFM{} discrete & \phantom{0}\underline{4.50 ± 0.76} & \textbf{97.33 ± 0.93} & \underline{30.542 ± 1.540} \\
   \oursrow
 & \CTFM{} cont. & \phantom{0}\textbf{4.38 ± 0.48} & \underline{97.31 ± 0.38} & \textbf{30.809 ± 1.099} \\
\midrule
\multirow[t]{4}{*}{Lumiere} & LCI & \phantom{0}8.38 & 88.35 & 21.631 \\
\cline{2-5}
& ConvLSTM & 34.79 ± 0.67 & \phantom{0}9.21 ± 2.81 & \phantom{0}9.217 ± 0.171 \\
 & SimVP & 71.03 ± 0.89 & \phantom{,}-1.92 ± 0.51 & \phantom{0}2.989 ± 0.109 \\
 & ViViT$^*$ & \phantom{000}$\mathrm{OOM}$ & \phantom{000}$\mathrm{OOM}$ & \phantom{000}$\mathrm{OOM}$ \\
  & NODE+LSTM & 13.07 ± 1.03 & 48.66 ± 2.26 & 17.742 ± 0.659 \\
  \oursrow
 &\CTFM{} discrete & \phantom{0}\underline{7.92 ± 0.92} & \textbf{91.43 ± 1.84} & \underline{22.427 ± 0.969} \\
 \oursrow
& \CTFM{} cont.  & \phantom{0}\textbf{7.55 ± 0.86} & \underline{89.32 ± 1.83} & \textbf{22.551 ± 0.979} \\

\bottomrule
\end{tabular}

\end{table}

\textbf{Baselines}
We compare \CTFM{} against established spatio-temporal learning methods.
As a clinically motivated heuristic, the \ib{} baseline (\IB{}) simply reuses the last available image and serves as a lower bound.
Among sequence models, we include ConvLSTM~\citep{shi_convolutional_2015}, SimVP~\citep{gao_simvp_2022}, and ViViT~\citep{arnab_vivit_2021} as representative recurrent, convolutional, and transformer backbones.
For continuous-time sequence modeling, we further evaluate an ODE-LSTM~\citep{lechner_learning_2020} baseline.
For the flow matching library we use ~\cite{tong_simulation-free_2024,tong_improving_2024,tong_torchcfm_2025}.
Together, these methods provide a spectrum of spatio-temporal architectures against which we benchmark \CTFM{}.
Computational requirements are described in detail in the appendix.
\\
\noindent
\textbf{Continuous vs. Discrete.}
We report results in two regimes: an \emph{discrete} setting, which allows direct comparison to existing spatio-temporal baselines, and a \emph{continuous} setting on ACDC, designed as an ablation to test the benefit of explicit timestamp conditioning.

\section{Results and Discussion}

\subsection{Towards unified benchmarking for medical 3D sequence-to-image forecasting}
We are among the first to propose an experimental setup for the sequence-to-image task, evaluating \CTFM{} under two complementary regimes.
The first uses \emph{discrete} input sequences, where some context images are missing but but acquisitions lie on a regular grid.
This setting enables comparison against established spatio-temporal baselines.
The second uses ACDC with resampled acquisitions to mimic \emph{continuous} input, allowing us to assess the benefit of explicit timestamp conditioning.
For completeness, we include an image-to-image (\textit{not sequence-to-image}) diffusion baseline on ACDC (details in ~\ref{subsec:image_to_image})
This required a two-stage training setup, first pretraining an autoencoder and then training the diffusion module for 1000 denoising steps, which already made the approach far more computationally demanding than all other baselines.
Iterative denoising leads to an order-of-magnitude longer inference time for a single image-to-image step and several orders of magnitude higher training cost, while not surpassing the simple \IB{} heuristic. 
\footnote{On our setup, a naive auto-regressive image-to-image \emph{latent-diffusion} pipeline applied across 11 context times per subject requires $\sim$5--6 hours \textit{per validation step}; see Appendix~\ref{sec:appendix} for details.}

\subsection{\CTFM{} is state-of-the-art for spatio-temporal 3D medical image forecasting}
\begin{wraptable}{r}{0.55\textwidth} 
\centering
\begin{tabular}{lccc}
\toprule
Method &  SSIM $\uparrow$ & PSNR $\uparrow$ & NRMSE $\downarrow$ \\
\midrule
LCI  & 93.27 & 29.77 & 0.0349 \\
\midrule
NODE + LSTM              & 57.50 & 22.87 & 0.0728 \\
\CTFM{} discr.              & 93.27 & 29.77 & 0.0348 \\
\oursrow
\CTFM{} cont.            & 93.86 & 30.09 & 0.0330 \\
\bottomrule
\end{tabular}
\caption{\textbf{Continuous ACDC}, where discrete \CTFM{} lacks explicit timestamp conditioning, and therefore fails to outperform \IB{}. Additional experiments in ~\ref{subsec:ablations} and in Table ~\ref{tab:break_continuous}}
\label{tab:val_loft_tfm_lci}
\end{wraptable}
Table~\ref{tab:main_results} reports the quantitative results across all three datasets.
We observe that both variants of \CTFM{} \textbf{substantially outperform} the competing spatio-temporal baselines, as well as \IB{}.
We also note that individually, \CTFM{} is better than \IB{} on each individual validation run.
On LUMIERE, which is characterized by very sparse and heterogeneous tumor trajectories, it is surprising that \CTFM{} is even able to outperform \IB{}.
These results demonstrate that \CTFM{} is effective across different temporal regimes: the discrete formulation already yields strong performance, while the continuous formulation provides further gains when timestampts are informative.
\CTFM{} runs \textbf{within the same computational budget} during inference (see Figure~\ref{fig:efficiency}) and in similar orders of magnitude (VRAM and wall-clock time) during training as natural imaging baselines (see~\ref{tab:runtime_memory}).
Further ablations are provided in ~\ref{sec:appendix}, confirming that \CTFM{} is stable across variations in \textit{feature size, training noise, and integration settings}.
While small differences appear, they are not substantial, indicating that our network is \textit{highly robust} to hyperparameter choices.

\subsection{\CTFM{} enables efficient flow-based continuous medical modeling}
Table~\ref{tab:val_loft_tfm_lci} demonstrates that \textit{incorporating explicit time embeddings improves forecasting} quality when scans occur at irregular intervals.
This shows that the continuous formulation of \CTFM{} is not only feasible but also beneficial in realistic clinical settings, where images are often irregularly sampled.
In fact, if we fully remove the timestamp information entirely, performance differences increase significantly, and the continuous variant clearly outperforms the discrete one~\ref{tab:break_continuous}.
Together, these results highlight that modeling real-valued timestamps can provide a measurable advantage over treating sequences as grid-aligned.
However, in Table~\ref{tab:main_results}, we see that using the discrete variant remains highly competitive.
Although any irregular series can in principle be quantized to a grid via $\mathcal{E}^{\mathrm{grid}}_{\mathbf g}$, doing so without loss requires increasingly fine grids. 
This becomes computationally inefficient, whereas the continuous variant scales with the number of context images and \textit{not} with the grid range $K\cdot\Delta$ (see~\eqref{eq:grid_definition}). 
This is reflected in Table~\ref{tab:runtime_memory} and Figure~\ref{fig:efficiency}, where continuous \CTFM{} is both more memory-efficient and faster to train than the discrete formulation.
It also highlights a broader limitation of the field: the scarcity of diverse spatio-temporal datasets in which real timing information is critical.

\subsection{\CTFM{} produces sharper reconstructions wit lower residuals}
\begin{figure}
    \centering
    \includegraphics[width=0.75\linewidth]{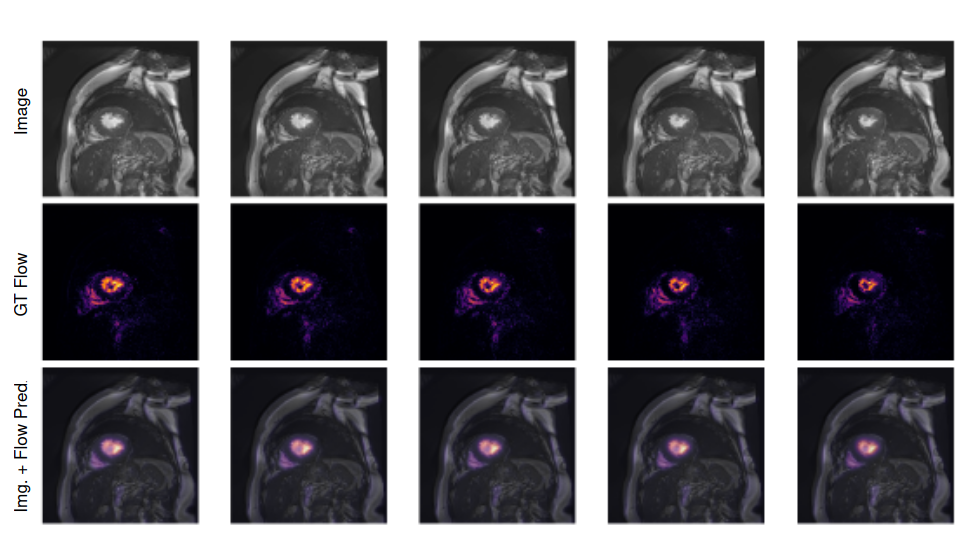}
    \caption{\textbf{Network Flows}: Top: input images at the first five timestamps. Middle: ground-truth voxel-wise differences ($|I_i - I_\text{target}|$. Bottom: predicted velocity fields $v_\theta(X_0,0)$, overlaid on the corresponding inputs. 
    The highlighted regions coincide with the areas of the largest temporal changes (primarily the ventricular cavities and myocardial boundaries).}
    \label{fig:placeholder}
\end{figure}
\begin{wrapfigure}{r}{0.57\linewidth}
    \centering
    \includegraphics[width=1\linewidth]{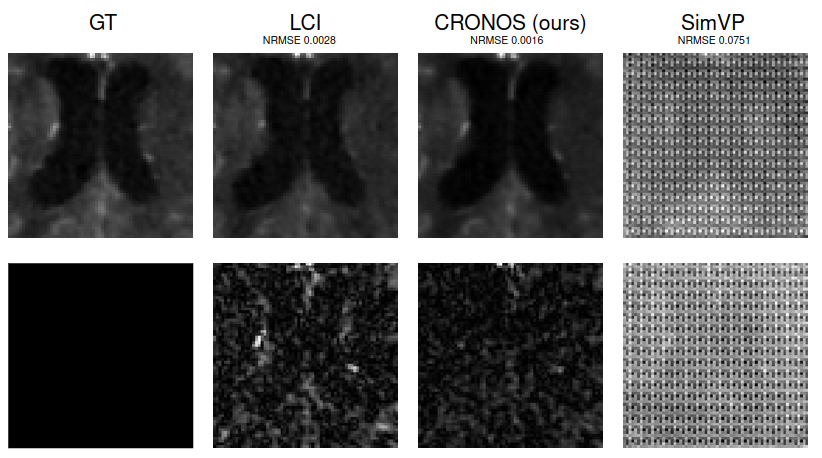}
    \caption{\textbf{Zoomed-in qualitative comparison on the ISLES dataset}. Ground truth (GT), Last Context Image (LCI), our method (\CTFM{}), and SimVP baseline. 
Shown here for visibility is a zoomed in patch of the qualitative results of the ISLES dataset.  }
    \label{fig:isles_qualitative}
\end{wrapfigure}

Figures~\ref{fig:qualitative_acdc},~\ref{fig:isles_qualitative} and ~\ref{fig:lumiere_qualitative} highlight qualitative comparisons, as well as dataset examples. 
The \IB{} baseline often appears visually close to the target, largely because many longitudinal scans exhibit only subtle changes.
However, e.g. SimVP tends to introduce artifacts and blur anatomical details.
In contrast, \CTFM{} yields sharper reconstructions and consistently lower residuals compared to \IB{}, highlighting its ability to capture fine-grained temporal progression.

\subsection{Future work: unlocking general spatio-temporal medical forecasting} 
While voxel-wise fidelity metrics such as NRMSE, PSNR, and SSIM remain the community standard, they do not fully capture clinically relevant trajectory modeling. 
As highlighted in recent efforts on image analysis validation ~\cite{maier-hein_metrics_2024}, such metrics may not always align with actual domain interest.
Developing metrics for spatio-temporal forecasting is therefore an important future direction.
In parallel, the scarcity of longitudinal and spatio-temporal datasets (beyond the ones we used in this study), poses a broader challenge for robust evaluation. 
Encouragingly, our results on LUMIERE suggest that progress is possible even under severe data limitations, and we hope to motivate further work on curating larger and more diverse publicly available cohorts.
Finally, the absence of large-scale foundation models for medical imaging, particularly in the spatio-temporal domain, remains a major bottleneck.
We view our work as a keystone contribution: establishing a unified flow-based framework for continuous spatio-temporal medical volumetric forecasting that can  \textit{both benefit from, and motivate}, future developments in medical imaging.



\section{Conclusion}
In this work, we presented \CTFM{} (\ctfm{}), a unified  spatio-temporal framework that forecasts 3D medical volumes at arbitrary target times by combining multiple context scans with  explicit real-valued time conditioning.
Unlike  single-image  or time-agnostic methods, \CTFM{} handles both grid-aligned and continuous timestamps within one architecture, and makes no disease-specific assumptions, it is among the first methods to demonstrate continuous sequence-to-image forecasting for 4D medical data.
Across three publicly available datasets (Cine-MRI, perfusion CT, longitudinal MRI), it outperforms baselines-including the strong \ib{} (\IB{})-and remains robust under hyperparameter changes while remaining computationally competitive.
By resolving the aforementioned limitations, our method enables clinically specific studies and advances patient-level forecasting for personalized precision medicine.

\section*{Broader Impact}
Longitudinal modeling of medical images has the potential to improve patient care by enabling earlier detection of disease progression, monitoring of treatment response, and improved personalization of therapy. By explicitly modeling continuous temporal evolution, our approach could support clinicians in making more informed decisions.
However, there are also risks: mispredictions may lead to incorrect clinical conclusions if models are deployed without careful validation and without a human in the loop.
Biases in training data (e.g., underrepresentation of certain populations or imaging modalities) may propagate to predictions, raising concerns about fairness and generalizability, which is a common problem in medical imaging.
We emphasize that our method is a research contribution intended to advance especially technical methodology. 
Clinical deployment would require extensive validation, regulatory approval, and integration into existing workflows. We believe that by releasing code and benchmarks, this work will support the community in building transparent, reproducible, and safe spatio-temporal models for healthcare.
But by proposing this method, we hope to support a general-purpose foundation for medical spatio-temporal and longitudinal modeling, which could massively propel this area forward.

\section*{Acknowledgments}
The present contribution is supported by the Helmholtz Association under the joint research school “HIDSS4Health – Helmholtz Information and Data Science School for Health.
\newpage

\bibliography{references}

\begin{thebibliography}{39}
\providecommand{\natexlab}[1]{#1}
\providecommand{\url}[1]{\texttt{#1}}
\expandafter\ifx\csname urlstyle\endcsname\relax
  \providecommand{\doi}[1]{doi: #1}\else
  \providecommand{\doi}{doi: \begingroup \urlstyle{rm}\Url}\fi

\bibitem[Arnab et~al.(2021)Arnab, Dehghani, Heigold, Sun, Lučić, and Schmid]{arnab_vivit_2021}
Anurag Arnab, Mostafa Dehghani, Georg Heigold, Chen Sun, Mario Lučić, and Cordelia Schmid.
\newblock {ViViT}: {A} {Video} {Vision} {Transformer}.
\newblock pp.\  6836--6846, 2021.
\newblock URL \url{https://openaccess.thecvf.com/content/ICCV2021/html/Arnab_ViViT_A_Video_Vision_Transformer_ICCV_2021_paper.html?ref=https://githubhelp.com}.

\bibitem[Bai \& Hong(2024)Bai and Hong]{bai_noder_2024}
Hao Bai and Yi~Hong.
\newblock {NODER}: {Image} {Sequence} {Regression} {Based} on {Neural} {Ordinary} {Differential} {Equations}, July 2024.
\newblock URL \url{http://arxiv.org/abs/2407.13241}.
\newblock arXiv:2407.13241 [cs].

\bibitem[Bernard et~al.(2018)Bernard, Lalande, Zotti, Cervenansky, Yang, Heng, Cetin, Lekadir, Camara, Gonzalez~Ballester, Sanroma, Napel, Petersen, Tziritas, Grinias, Khened, Kollerathu, Krishnamurthi, Rohé, Pennec, Sermesant, Isensee, Jäger, Maier-Hein, Full, Wolf, Engelhardt, Baumgartner, Koch, Wolterink, Išgum, Jang, Hong, Patravali, Jain, Humbert, and Jodoin]{bernard_deep_2018}
Olivier Bernard, Alain Lalande, Clement Zotti, Frederick Cervenansky, Xin Yang, Pheng-Ann Heng, Irem Cetin, Karim Lekadir, Oscar Camara, Miguel~Angel Gonzalez~Ballester, Gerard Sanroma, Sandy Napel, Steffen Petersen, Georgios Tziritas, Elias Grinias, Mahendra Khened, Varghese~Alex Kollerathu, Ganapathy Krishnamurthi, Marc-Michel Rohé, Xavier Pennec, Maxime Sermesant, Fabian Isensee, Paul Jäger, Klaus~H. Maier-Hein, Peter~M. Full, Ivo Wolf, Sandy Engelhardt, Christian~F. Baumgartner, Lisa~M. Koch, Jelmer~M. Wolterink, Ivana Išgum, Yeonggul Jang, Yoonmi Hong, Jay Patravali, Shubham Jain, Olivier Humbert, and Pierre-Marc Jodoin.
\newblock Deep {Learning} {Techniques} for {Automatic} {MRI} {Cardiac} {Multi}-{Structures} {Segmentation} and {Diagnosis}: {Is} the {Problem} {Solved}?
\newblock \emph{IEEE Transactions on Medical Imaging}, 37\penalty0 (11):\penalty0 2514--2525, November 2018.
\newblock ISSN 0278-0062, 1558-254X.
\newblock \doi{10.1109/TMI.2018.2837502}.
\newblock URL \url{https://ieeexplore.ieee.org/document/8360453/}.

\bibitem[Chen et~al.(2025)Chen, Zhang, Han, Wen, and Yu]{chen_reflections_2025}
Durong Chen, Meiling Zhang, Hongjuan Han, Yalu Wen, and Hongmei Yu.
\newblock Reflections on dynamic prediction of {Alzheimer}’s disease: advancements in modeling longitudinal outcomes and time-to-event data.
\newblock \emph{BMC Medical Research Methodology}, 25\penalty0 (1):\penalty0 175, July 2025.
\newblock ISSN 1471-2288.
\newblock \doi{10.1186/s12874-025-02618-x}.
\newblock URL \url{https://doi.org/10.1186/s12874-025-02618-x}.

\bibitem[Chen et~al.(2019)Chen, Rubanova, Bettencourt, and Duvenaud]{chen_neural_2019}
Ricky T.~Q. Chen, Yulia Rubanova, Jesse Bettencourt, and David Duvenaud.
\newblock Neural {Ordinary} {Differential} {Equations}, December 2019.
\newblock URL \url{http://arxiv.org/abs/1806.07366}.
\newblock arXiv:1806.07366 [cs].

\bibitem[Fang et~al.(2021)Fang, Bai, Chen, Zhou, Xia, Qin, Gong, Xie, Zhou, Tu, Zhang, Liu, Chen, Bai, and Torr]{fang_deep_2021}
Cong Fang, Song Bai, Qianlan Chen, Yu~Zhou, Liming Xia, Lixin Qin, Shi Gong, Xudong Xie, Chunhua Zhou, Dandan Tu, Changzheng Zhang, Xiaowu Liu, Weiwei Chen, Xiang Bai, and Philip H.~S. Torr.
\newblock Deep learning for predicting {COVID}-19 malignant progression.
\newblock \emph{Medical Image Analysis}, 72:\penalty0 102096, August 2021.
\newblock ISSN 1361-8415.
\newblock \doi{10.1016/j.media.2021.102096}.
\newblock URL \url{https://www.sciencedirect.com/science/article/pii/S1361841521001420}.

\bibitem[Farhad et~al.(2023)Farhad, Masud, Beg, Ahmad, and Ahmed]{farhad_review_2023}
Moomal Farhad, Mohammad~Mehedy Masud, Azam Beg, Amir Ahmad, and Luai Ahmed.
\newblock A {Review} of {Medical} {Diagnostic} {Video} {Analysis} {Using} {Deep} {Learning} {Techniques}.
\newblock \emph{Applied Sciences}, 13\penalty0 (11):\penalty0 6582, January 2023.
\newblock ISSN 2076-3417.
\newblock \doi{10.3390/app13116582}.
\newblock URL \url{https://www.mdpi.com/2076-3417/13/11/6582}.
\newblock Publisher: Multidisciplinary Digital Publishing Institute.

\bibitem[Gao et~al.(2022)Gao, Tan, Wu, and Li]{gao_simvp_2022}
Zhangyang Gao, Cheng Tan, Lirong Wu, and Stan~Z. Li.
\newblock {SimVP}: {Simpler} {Yet} {Better} {Video} {Prediction}.
\newblock pp.\  3170--3180, 2022.
\newblock URL \url{https://openaccess.thecvf.com/content/CVPR2022/html/Gao_SimVP_Simpler_Yet_Better_Video_Prediction_CVPR_2022_paper.html}.

\bibitem[Gomes et~al.(2022)Gomes, Vwalika, Lee, Willis, Sieniek, Price, Chen, Kasaro, Taylor, Stringer, McKinney, Sindano, Dahl, III, Gilmer, Chi, Lau, Spitz, Saensuksopa, Liu, Wong, Pilgrim, Uddin, Corrado, Peng, Chou, Tse, Stringer, and Shetty]{gomes_ai_2022}
Ryan~G. Gomes, Bellington Vwalika, Chace Lee, Angelica Willis, Marcin Sieniek, Joan~T. Price, Christina Chen, Margaret~P. Kasaro, James~A. Taylor, Elizabeth~M. Stringer, Scott~Mayer McKinney, Ntazana Sindano, George~E. Dahl, William~Goodnight III, Justin Gilmer, Benjamin~H. Chi, Charles Lau, Terry Spitz, T.~Saensuksopa, Kris Liu, Jonny Wong, Rory Pilgrim, Akib Uddin, Greg Corrado, Lily Peng, Katherine Chou, Daniel Tse, Jeffrey S.~A. Stringer, and Shravya Shetty.
\newblock {AI} system for fetal ultrasound in low-resource settings, March 2022.
\newblock URL \url{http://arxiv.org/abs/2203.10139}.
\newblock arXiv:2203.10139 [cs].

\bibitem[Lachinov et~al.(2022)Lachinov, Chakravarty, Grechenig, Schmidt-Erfurth, and Bogunovic]{lachinov_learning_2022}
Dmitrii Lachinov, Arunava Chakravarty, Christoph Grechenig, Ursula Schmidt-Erfurth, and Hrvoje Bogunovic.
\newblock Learning {Spatio}-{Temporal} {Model} of {Disease} {Progression} with {NeuralODEs} from {Longitudinal} {Volumetric} {Data}, November 2022.
\newblock URL \url{http://arxiv.org/abs/2211.04234}.
\newblock arXiv:2211.04234 [cs].

\bibitem[Lechner \& Hasani(2020)Lechner and Hasani]{lechner_learning_2020}
Mathias Lechner and Ramin~M. Hasani.
\newblock Learning {Long}-{Term} {Dependencies} in {Irregularly}-{Sampled} {Time} {Series}.
\newblock \emph{ArXiv}, June 2020.
\newblock URL \url{https://www.semanticscholar.org/paper/Learning-Long-Term-Dependencies-in-Time-Series-Lechner-Hasani/4d9521fbd135559e4d186e96b703f3bd8fd7617e}.

\bibitem[Li et~al.(2024)Li, Zhao, Li, and Li]{li_deep_2024}
Yunlong Li, Zijian Zhao, Renbo Li, and Feng Li.
\newblock Deep learning for surgical workflow analysis: a survey of progresses, limitations, and trends.
\newblock \emph{Artificial Intelligence Review}, 57\penalty0 (11):\penalty0 291, September 2024.
\newblock ISSN 1573-7462.
\newblock \doi{10.1007/s10462-024-10929-6}.
\newblock URL \url{https://doi.org/10.1007/s10462-024-10929-6}.

\bibitem[Lipman et~al.(2023)Lipman, Chen, Ben-Hamu, Nickel, and Le]{lipman_flow_2023}
Yaron Lipman, Ricky T.~Q. Chen, Heli Ben-Hamu, Maximilian Nickel, and Matt Le.
\newblock Flow {Matching} for {Generative} {Modeling}, February 2023.
\newblock URL \url{http://arxiv.org/abs/2210.02747}.
\newblock arXiv:2210.02747 [cs].

\bibitem[Lipman et~al.(2024)Lipman, Havasi, Holderrieth, Shaul, Le, Karrer, Chen, Lopez-Paz, Ben-Hamu, and Gat]{lipman_flow_2024}
Yaron Lipman, Marton Havasi, Peter Holderrieth, Neta Shaul, Matt Le, Brian Karrer, Ricky T.~Q. Chen, David Lopez-Paz, Heli Ben-Hamu, and Itai Gat.
\newblock Flow {Matching} {Guide} and {Code}, December 2024.
\newblock URL \url{http://arxiv.org/abs/2412.06264}.
\newblock arXiv:2412.06264 [cs].

\bibitem[Litrico et~al.(2024)Litrico, Guarnera, Giuffrida, Ravì, and Battiato]{linguraru_tadm_2024}
Mattia Litrico, Francesco Guarnera, Mario~Valerio Giuffrida, Daniele Ravì, and Sebastiano Battiato.
\newblock {TADM}: {Temporally}-{Aware} {Diffusion} {Model} for {Neurodegenerative} {Progression} on {Brain} {MRI}.
\newblock In Marius~George Linguraru, Qi~Dou, Aasa Feragen, Stamatia Giannarou, Ben Glocker, Karim Lekadir, and Julia~A. Schnabel (eds.), \emph{Medical {Image} {Computing} and {Computer} {Assisted} {Intervention} – {MICCAI} 2024}, volume 15002, pp.\  444--453. Springer Nature Switzerland, Cham, 2024.
\newblock ISBN 978-3-031-72068-0 978-3-031-72069-7.
\newblock \doi{10.1007/978-3-031-72069-7_42}.
\newblock URL \url{https://link.springer.com/10.1007/978-3-031-72069-7_42}.
\newblock Series Title: Lecture Notes in Computer Science.

\bibitem[Liu et~al.(2025)Liu, Xu, Shen, Huguet, Wang, Tong, Bzdok, Stewart, Wang, Priore, and Krishnaswamy]{liu_imageflownet_2025}
Chen Liu, Ke~Xu, Liangbo~L. Shen, Guillaume Huguet, Zilong Wang, Alexander Tong, Danilo Bzdok, Jay Stewart, Jay~C. Wang, Lucian V.~Del Priore, and Smita Krishnaswamy.
\newblock {ImageFlowNet}: {Forecasting} {Multiscale} {Image}-{Level} {Trajectories} of {Disease} {Progression} with {Irregularly}-{Sampled} {Longitudinal} {Medical} {Images}, April 2025.
\newblock URL \url{http://arxiv.org/abs/2406.14794}.
\newblock arXiv:2406.14794 [eess].

\bibitem[Maier-Hein et~al.(2024)Maier-Hein, Reinke, Godau, Tizabi, Buettner, Christodoulou, Glocker, Isensee, Kleesiek, Kozubek, Reyes, Riegler, Wiesenfarth, Kavur, Sudre, Baumgartner, Eisenmann, Heckmann-Nötzel, Rädsch, Acion, Antonelli, Arbel, Bakas, Benis, Blaschko, Cardoso, Cheplygina, Cimini, Collins, Farahani, Ferrer, Galdran, van Ginneken, Haase, Hashimoto, Hoffman, Huisman, Jannin, Kahn, Kainmueller, Kainz, Karargyris, Karthikesalingam, Kofler, Kopp-Schneider, Kreshuk, Kurc, Landman, Litjens, Madani, Maier-Hein, Martel, Mattson, Meijering, Menze, Moons, Müller, Nichyporuk, Nickel, Petersen, Rajpoot, Rieke, Saez-Rodriguez, Sánchez, Shetty, van Smeden, Summers, Taha, Tiulpin, Tsaftaris, Van~Calster, Varoquaux, and Jäger]{maier-hein_metrics_2024}
Lena Maier-Hein, Annika Reinke, Patrick Godau, Minu~D. Tizabi, Florian Buettner, Evangelia Christodoulou, Ben Glocker, Fabian Isensee, Jens Kleesiek, Michal Kozubek, Mauricio Reyes, Michael~A. Riegler, Manuel Wiesenfarth, A.~Emre Kavur, Carole~H. Sudre, Michael Baumgartner, Matthias Eisenmann, Doreen Heckmann-Nötzel, Tim Rädsch, Laura Acion, Michela Antonelli, Tal Arbel, Spyridon Bakas, Arriel Benis, Matthew~B. Blaschko, M.~Jorge Cardoso, Veronika Cheplygina, Beth~A. Cimini, Gary~S. Collins, Keyvan Farahani, Luciana Ferrer, Adrian Galdran, Bram van Ginneken, Robert Haase, Daniel~A. Hashimoto, Michael~M. Hoffman, Merel Huisman, Pierre Jannin, Charles~E. Kahn, Dagmar Kainmueller, Bernhard Kainz, Alexandros Karargyris, Alan Karthikesalingam, Florian Kofler, Annette Kopp-Schneider, Anna Kreshuk, Tahsin Kurc, Bennett~A. Landman, Geert Litjens, Amin Madani, Klaus Maier-Hein, Anne~L. Martel, Peter Mattson, Erik Meijering, Bjoern Menze, Karel G.~M. Moons, Henning Müller, Brennan Nichyporuk, Felix Nickel, Jens
  Petersen, Nasir Rajpoot, Nicola Rieke, Julio Saez-Rodriguez, Clara~I. Sánchez, Shravya Shetty, Maarten van Smeden, Ronald~M. Summers, Abdel~A. Taha, Aleksei Tiulpin, Sotirios~A. Tsaftaris, Ben Van~Calster, Gaël Varoquaux, and Paul~F. Jäger.
\newblock Metrics reloaded: recommendations for image analysis validation.
\newblock \emph{Nature Methods}, 21\penalty0 (2):\penalty0 195--212, February 2024.
\newblock ISSN 1548-7105.
\newblock \doi{10.1038/s41592-023-02151-z}.
\newblock URL \url{https://www.nature.com/articles/s41592-023-02151-z}.
\newblock Publisher: Nature Publishing Group.

\bibitem[Martí-Juan et~al.(2020)Martí-Juan, Sanroma-Guell, and Piella]{marti-juan_survey_2020}
Gerard Martí-Juan, Gerard Sanroma-Guell, and Gemma Piella.
\newblock A survey on machine and statistical learning for longitudinal analysis of neuroimaging data in {Alzheimer}’s disease.
\newblock \emph{Computer Methods and Programs in Biomedicine}, 189:\penalty0 105348, June 2020.
\newblock ISSN 0169-2607.
\newblock \doi{10.1016/j.cmpb.2020.105348}.
\newblock URL \url{https://www.sciencedirect.com/science/article/pii/S0169260719316165}.

\bibitem[Park et~al.(2021)Park, Kim, Lee, Choo, Lee, Kim, and Choi]{park_vid-ode_2021}
Sunghyun Park, Kangyeol Kim, Junsoo Lee, Jaegul Choo, Joonseok Lee, Sookyung Kim, and Edward Choi.
\newblock Vid-{ODE}: {Continuous}-{Time} {Video} {Generation} with {Neural} {Ordinary} {Differential} {Equation}, March 2021.
\newblock URL \url{http://arxiv.org/abs/2010.08188}.
\newblock arXiv:2010.08188 [cs].

\bibitem[Petersen et~al.(2010)Petersen, Aisen, Beckett, Donohue, Gamst, Harvey, Jack, Jagust, Shaw, Toga, Trojanowski, and Weiner]{petersen_alzheimers_2010}
R~C. Petersen, P~S. Aisen, L~A. Beckett, M~C. Donohue, A~C. Gamst, D~J. Harvey, C~R. Jack, W~J. Jagust, L~M. Shaw, A~W. Toga, J~Q. Trojanowski, and M~W. Weiner.
\newblock Alzheimer's {Disease} {Neuroimaging} {Initiative} ({ADNI}).
\newblock \emph{Neurology}, 74\penalty0 (3):\penalty0 201--209, January 2010.
\newblock ISSN 0028-3878.
\newblock \doi{10.1212/WNL.0b013e3181cb3e25}.
\newblock URL \url{https://www.ncbi.nlm.nih.gov/pmc/articles/PMC2809036/}.

\bibitem[Puglisi et~al.(2025)Puglisi, Alexander, and Ravì]{puglisi_brain_2025}
Lemuel Puglisi, Daniel~C. Alexander, and Daniele Ravì.
\newblock Brain {Latent} {Progression}: {Individual}-based {Spatiotemporal} {Disease} {Progression} on {3D} {Brain} {MRIs} via {Latent} {Diffusion}, February 2025.
\newblock URL \url{http://arxiv.org/abs/2502.08560}.
\newblock arXiv:2502.08560 [cs].

\bibitem[Riedel et~al.(2024)Riedel, de~la Rosa, Baran, Petzsche, Baazaoui, Yang, Robben, Seia, Wiest, Reyes, Su, Zimmer, Boeckh-Behrens, Berndt, Menze, Wiestler, Wegener, and Kirschke]{riedel_isles_2024}
Evamaria~O. Riedel, Ezequiel de~la Rosa, The~Anh Baran, Moritz~Hernandez Petzsche, Hakim Baazaoui, Kaiyuan Yang, David Robben, Joaquin~Oscar Seia, Roland Wiest, Mauricio Reyes, Ruisheng Su, Claus Zimmer, Tobias Boeckh-Behrens, Maria Berndt, Bjoern Menze, Benedikt Wiestler, Susanne Wegener, and Jan~S. Kirschke.
\newblock {ISLES} 2024: {The} first longitudinal multimodal multi-center real-world dataset in (sub-)acute stroke.
\newblock 2024.
\newblock \doi{10.48550/ARXIV.2408.11142}.
\newblock URL \url{https://arxiv.org/abs/2408.11142}.
\newblock Publisher: arXiv Version Number: 1.

\bibitem[Rivail et~al.(2019)Rivail, Schmidt-Erfurth, Vogl, Waldstein, Riedl, Grechenig, Wu, and Bogunović]{rivail_modeling_2019}
Antoine Rivail, Ursula Schmidt-Erfurth, Wolf-Dieter Vogl, Sebastian~M. Waldstein, Sophie Riedl, Christoph Grechenig, Zhichao Wu, and Hrvoje Bogunović.
\newblock Modeling {Disease} {Progression} {In} {Retinal} {OCTs} {With} {Longitudinal} {Self}-{Supervised} {Learning}, October 2019.
\newblock URL \url{http://arxiv.org/abs/1910.09420}.
\newblock arXiv:1910.09420 [eess].

\bibitem[Rombach et~al.(2022)Rombach, Blattmann, Lorenz, Esser, and Ommer]{rombach_high-resolution_2022}
Robin Rombach, Andreas Blattmann, Dominik Lorenz, Patrick Esser, and Bjorn Ommer.
\newblock High-{Resolution} {Image} {Synthesis} with {Latent} {Diffusion} {Models}.
\newblock \emph{2022 IEEE/CVF Conference on Computer Vision and Pattern Recognition (CVPR)}, pp.\  10674--10685, June 2022.
\newblock \doi{10.1109/CVPR52688.2022.01042}.
\newblock URL \url{https://ieeexplore.ieee.org/document/9878449/}.
\newblock Conference Name: 2022 IEEE/CVF Conference on Computer Vision and Pattern Recognition (CVPR) ISBN: 9781665469463 Place: New Orleans, LA, USA Publisher: IEEE.

\bibitem[SHI et~al.(2015)SHI, Chen, Wang, Yeung, Wong, and WOO]{shi_convolutional_2015}
Xingjian SHI, Zhourong Chen, Hao Wang, Dit-Yan Yeung, Wai-kin Wong, and Wang-chun WOO.
\newblock Convolutional {LSTM} {Network}: {A} {Machine} {Learning} {Approach} for {Precipitation} {Nowcasting}.
\newblock In \emph{Advances in {Neural} {Information} {Processing} {Systems}}, volume~28. Curran Associates, Inc., 2015.
\newblock URL \url{https://proceedings.neurips.cc/paper/2015/hash/07563a3fe3bbe7e3ba84431ad9d055af-Abstract.html}.

\bibitem[Suter et~al.(2022)Suter, Knecht, Valenzuela, Notter, Hewer, Schucht, Wiest, and Reyes]{suter_lumiere_2022}
Yannick Suter, Urspeter Knecht, Waldo Valenzuela, Michelle Notter, Ekkehard Hewer, Philippe Schucht, Roland Wiest, and Mauricio Reyes.
\newblock The {LUMIERE} dataset: {Longitudinal} {Glioblastoma} {MRI} with expert {RANO} evaluation.
\newblock \emph{Scientific Data}, 9\penalty0 (1):\penalty0 768, December 2022.
\newblock ISSN 2052-4463.
\newblock \doi{10.1038/s41597-022-01881-7}.
\newblock URL \url{https://www.nature.com/articles/s41597-022-01881-7}.
\newblock Publisher: Nature Publishing Group.

\bibitem[Tancik et~al.(2020)Tancik, Srinivasan, Mildenhall, Fridovich-Keil, Raghavan, Singhal, Ramamoorthi, Barron, and Ng]{tancik_fourier_2020}
Matthew Tancik, Pratul~P. Srinivasan, B.~Mildenhall, Sara Fridovich-Keil, Nithin Raghavan, Utkarsh Singhal, R.~Ramamoorthi, J.~Barron, and Ren Ng.
\newblock Fourier {Features} {Let} {Networks} {Learn} {High} {Frequency} {Functions} in {Low} {Dimensional} {Domains}.
\newblock \emph{ArXiv}, June 2020.
\newblock URL \url{https://www.semanticscholar.org/paper/Fourier-Features-Let-Networks-Learn-High-Frequency-Tancik-Srinivasan/a0dc3135c40e150f0271002a96b7c9680b6cac40}.

\bibitem[Therasse et~al.(2000)Therasse, Arbuck, Eisenhauer, Wanders, Kaplan, Rubinstein, Verweij, Van~Glabbeke, Van~Oosterom, Christian, and Gwyther]{therasse_new_2000}
Patrick Therasse, Susan~G. Arbuck, Elizabeth~A. Eisenhauer, Jantien Wanders, Richard~S. Kaplan, Larry Rubinstein, Jaap Verweij, Martine Van~Glabbeke, Allan~T. Van~Oosterom, Michaele~C. Christian, and Steve~G. Gwyther.
\newblock New {Guidelines} to {Evaluate} the {Response} to {Treatment} in {Solid} {Tumors}.
\newblock \emph{JNCI: Journal of the National Cancer Institute}, 92\penalty0 (3):\penalty0 205--216, February 2000.
\newblock ISSN 0027-8874, 1460-2105.
\newblock \doi{10.1093/jnci/92.3.205}.
\newblock URL \url{https://academic.oup.com/jnci/jnci/article/2965042/New}.
\newblock Publisher: Oxford University Press (OUP).

\bibitem[Tong(2025)]{tong_torchcfm_2025}
Alexander Tong.
\newblock {TorchCFM}, January 2025.
\newblock URL \url{https://github.com/atong01/conditional-flow-matching}.

\bibitem[Tong et~al.(2024{\natexlab{a}})Tong, Fatras, Malkin, Huguet, Zhang, Rector-Brooks, Wolf, and Bengio]{tong_improving_2024}
Alexander Tong, Kilian Fatras, Nikolay Malkin, Guillaume Huguet, Yanlei Zhang, Jarrid Rector-Brooks, Guy Wolf, and Yoshua Bengio.
\newblock Improving and generalizing flow-based generative models with minibatch optimal transport, March 2024{\natexlab{a}}.
\newblock URL \url{http://arxiv.org/abs/2302.00482}.
\newblock arXiv:2302.00482 [cs].

\bibitem[Tong et~al.(2024{\natexlab{b}})Tong, Malkin, Fatras, Atanackovic, Zhang, Huguet, Wolf, and Bengio]{tong_simulation-free_2024}
Alexander Tong, Nikolay Malkin, Kilian Fatras, Lazar Atanackovic, Yanlei Zhang, Guillaume Huguet, Guy Wolf, and Yoshua Bengio.
\newblock Simulation-free {Schrödinger} bridges via score and flow matching, March 2024{\natexlab{b}}.
\newblock URL \url{http://arxiv.org/abs/2307.03672}.
\newblock arXiv:2307.03672 [cs].

\bibitem[Voleti et~al.(2022)Voleti, Jolicoeur-Martineau, and Pal]{voleti_mcvd_2022}
Vikram Voleti, Alexia Jolicoeur-Martineau, and Christopher Pal.
\newblock {MCVD}: {Masked} {Conditional} {Video} {Diffusion} for {Prediction}, {Generation}, and {Interpolation}.
\newblock 2022.
\newblock \doi{10.48550/ARXIV.2205.09853}.
\newblock URL \url{https://arxiv.org/abs/2205.09853}.
\newblock Publisher: arXiv Version Number: 4.

\bibitem[Yan et~al.(2021)Yan, Zhang, Abbeel, and Srinivas]{yan_videogpt_2021}
Wilson Yan, Yunzhi Zhang, P.~Abbeel, and A.~Srinivas.
\newblock {VideoGPT}: {Video} {Generation} using {VQ}-{VAE} and {Transformers}.
\newblock \emph{ArXiv}, April 2021.
\newblock URL \url{https://www.semanticscholar.org/paper/VideoGPT%3A-Video-Generation-using-VQ-VAE-and-Yan-Zhang/2d9ae4c167510ed78803735fc57ea67c3cc55a35}.

\bibitem[Ye \& Bilodeau(2023)Ye and Bilodeau]{ye_stdiff_2023}
Xi~Ye and Guillaume-Alexandre Bilodeau.
\newblock {STDiff}: {Spatio}-temporal {Diffusion} for {Continuous} {Stochastic} {Video} {Prediction}.
\newblock 2023.
\newblock \doi{10.48550/ARXIV.2312.06486}.
\newblock URL \url{https://arxiv.org/abs/2312.06486}.
\newblock Publisher: arXiv Version Number: 1.

\bibitem[Yoon et~al.(2024)Yoon, Myong, Kim, Sim, Cho, Oh, and Kim]{yoon_latent_2024}
Dan Yoon, Youho Myong, Young~Gyun Kim, Yongsik Sim, Minwoo Cho, Byung-Mo Oh, and Sungwan Kim.
\newblock Latent diffusion model-based {MRI} superresolution enhances mild cognitive impairment prognostication and {Alzheimer}'s disease classification.
\newblock \emph{NeuroImage}, 296:\penalty0 120663, August 2024.
\newblock ISSN 1053-8119.
\newblock \doi{10.1016/j.neuroimage.2024.120663}.
\newblock URL \url{https://www.sciencedirect.com/science/article/pii/S1053811924001587}.

\bibitem[Zhang et~al.(2025{\natexlab{a}})Zhang, Chattopadhyay, Thomopoulos, Ambite, Thompson, and Steeg]{zhang_diffusion_2025}
Shaorong Zhang, Tamoghna Chattopadhyay, Sophia~I. Thomopoulos, Jose-Luis Ambite, Paul~M. Thompson, and Greg~Ver Steeg.
\newblock Diffusion {Bridge} {Models} for {3D} {Medical} {Image} {Translation}, April 2025{\natexlab{a}}.
\newblock URL \url{http://arxiv.org/abs/2504.15267}.
\newblock arXiv:2504.15267 [cs].

\bibitem[Zhang et~al.(2025{\natexlab{b}})Zhang, Pu, Kawamura, Loza, Bengio, Shung, and Tong]{zhang_trajectory_2025}
Xi~Zhang, Yuan Pu, Yuki Kawamura, Andrew Loza, Yoshua Bengio, Dennis~L. Shung, and Alexander Tong.
\newblock Trajectory {Flow} {Matching} with {Applications} to {Clinical} {Time} {Series} {Modeling}, February 2025{\natexlab{b}}.
\newblock URL \url{http://arxiv.org/abs/2410.21154}.
\newblock arXiv:2410.21154 [cs].

\bibitem[Zhu et~al.(2024{\natexlab{a}})Zhu, Tao, Tao, Deng, Cai, Wu, Wang, Tang, Zhu, Gu, Shen, and Zhang]{linguraru_loci-diffcom_2024}
Zihao Zhu, Tianli Tao, Yitian Tao, Haowen Deng, Xinyi Cai, Gaofeng Wu, Kaidong Wang, Haifeng Tang, Lixuan Zhu, Zhuoyang Gu, Dinggang Shen, and Han Zhang.
\newblock {LoCI}-{DiffCom}: {Longitudinal} {Consistency}-{Informed} {Diffusion} {Model} for {3D} {Infant} {Brain} {Image} {Completion}.
\newblock In Marius~George Linguraru, Qi~Dou, Aasa Feragen, Stamatia Giannarou, Ben Glocker, Karim Lekadir, and Julia~A. Schnabel (eds.), \emph{Medical {Image} {Computing} and {Computer} {Assisted} {Intervention} – {MICCAI} 2024}, volume 15002, pp.\  249--258. Springer Nature Switzerland, Cham, 2024{\natexlab{a}}.
\newblock ISBN 978-3-031-72068-0 978-3-031-72069-7.
\newblock \doi{10.1007/978-3-031-72069-7_24}.
\newblock URL \url{https://link.springer.com/10.1007/978-3-031-72069-7_24}.
\newblock Series Title: Lecture Notes in Computer Science.

\bibitem[Zhu et~al.(2024{\natexlab{b}})Zhu, Tao, Tao, Deng, Cai, Wu, Wang, Tang, Zhu, Gu, Shen, and Zhang]{zhu_loci-diffcom_2024}
Zihao Zhu, Tianli Tao, Yitian Tao, Haowen Deng, Xinyi Cai, Gaofeng Wu, Kaidong Wang, Haifeng Tang, Lixuan Zhu, Zhuoyang Gu, Dinggang Shen, and Han Zhang.
\newblock {LoCI}-{DiffCom}: {Longitudinal} {Consistency}-{Informed} {Diffusion} {Model} for {3D} {Infant} {Brain} {Image} {Completion}.
\newblock In Marius~George Linguraru, Qi~Dou, Aasa Feragen, Stamatia Giannarou, Ben Glocker, Karim Lekadir, and Julia~A. Schnabel (eds.), \emph{Medical {Image} {Computing} and {Computer} {Assisted} {Intervention} – {MICCAI} 2024}, pp.\  249--258, Cham, 2024{\natexlab{b}}. Springer Nature Switzerland.
\newblock ISBN 978-3-031-72069-7.
\newblock \doi{10.1007/978-3-031-72069-7_24}.

\end{thebibliography}
\bibliographystyle{iclr2026_conference}
\newpage
\appendix
\section{Appendix}\label{sec:appendix}

\subsection{Discrete \CTFM{} Details}\label{subsec:appendix_sparsity_filling}

\subsubsection{Grid Embedding}\label{app:grid_embedding}
Assume we want to embed data on a grid with spacing $\Delta > 0$, and a maximum size of $K\in \mathbb{N}_+$, where
\begin{equation}
    g_k = g_1 + (k-1)\Delta, \quad k\in \{1, \dots, K\}.
    \label{eq:grid_definition}
\end{equation}
In total, we define the grid as $\mathbf g=[g_1,\dots,g_k]$.
We define the grid quantizer as
\begin{equation}
    q(t_i) = \mathrm{clip}\!\left(1+\left\lfloor \frac{t_i-g_1}{\Delta}+\tfrac{1}{2}\right\rfloor,\,1,\,K\right),
\end{equation}
where $\mathrm{clip}(a,b,c) \coloneqq \mathrm{min}(\mathrm{max}(a,b),c)$.
I.e. we clip the value $t$ to the closest grid point $g_k$.
We define the Kronecker delta for indices
\begin{equation}
    \delta_{k, q(t_i)}=\begin{cases}
1, & q(t_i) = k\\
0, & \text{else}.
\end{cases}
\end{equation}
Since for some grid sizes $\Delta$, there can be too many items, we define the occupancy:
\begin{equation}\label{eq:occupancy_op}
    m_k = \mathrm{min} \left ( 1, \sum_{i=1}^T \delta_{k, q(t_i)}  \right ) \in \{0,1\}.
\end{equation}
We use the last available index for filling:
\begin{equation}
    i^*(k) = \mathrm{arg max}_i (\delta_{k, q(t_i)} t_i),
\end{equation}
i.e. the index which is \textit{latest} while still falling into the index $k$.
Finally, our embedded images are 
\begin{equation}
        \tilde I_k =  \begin{cases}
I_{i^*(k)}, & m_k=1\\
0, & \text{else}.
\end{cases}
\end{equation}
Lastly we define the grid embedding as 
\begin{equation}
    \mathcal{E}^{\mathrm{grid}}_{\mathbf g} \!\left(\{(I_i,t_i)\}_{i=1}^T\right) = [\tilde I_1, \dots, \tilde I_K].
    \label{eq:grid_embedding_def}
\end{equation}

\subsubsection{Last Observed Carry Forward.}  \label{app:sparsity_filling}
Longitudinal imaging sequences are often incomplete, with missing acquisitions at certain timestamps.  
To ensure consistent tensor inputs, we adopt the last-observed carry-forward procedure: missing frames are initialized as zero images and then replaced by the nearest available context scan in time.  
We iteratively define the 
\begin{equation}
    \hat I_1 = \tilde I_{k_0}, \quad \hat I_k = m_k \tilde I_k + (1-m_k) \hat I_{k-1} \quad k \in \{2,\dots,K\},
\end{equation}
where $k_0$ defines the first observed image in the grid
\begin{equation}
    k_0 = \mathrm{min} \{k\in\{1,\dots,K\}\}|m_k=1\}.
\end{equation}
We define shorthand: 
\begin{equation}
    \mathcal{F}^{\mathrm{LOCF}} \left (  [\tilde I_1, \dots, \tilde I_K] \right) =  [\hat I_1, \dots, \hat I_K].
\end{equation}
This guarantees that the model always observes a full sequence $\{I_1,\dots,I_T\}$ while still preserving the true irregularity of acquisition.  
Empirically, sparsity filling stabilizes training and improves performance compared to leaving missing slots empty, since it prevents the network from confusing acquisition gaps with valid image content.
This can be seen in the ablations, see Table ~\ref{tab:ablation_different_design_choices}.

\begin{algorithm*}
\caption{\CTFM{}: Discrete training and Inference}
\begin{algorithmic}[1]
  \Require Patients $\mathcal{P} = \{[\mathcal{I}_1 , I_{\text{target},1}], \dots, [\mathcal{I}_p , I_{\text{target},p}]\}$ and initial network $v_\theta$
  \While{training}
    \State $[\mathcal{I} , I_{\text{target}}] \sim \mathcal{P}(\mathcal{X})$ \Comment{pick a random patient}
    \State  $\tau \sim \mathcal{U}(0,1)$ \Comment{pick a random flow step}
    \State $\mathcal{I}_\text{target} \gets [I_\text{target}, \dots, I_\text{target}]$ \Comment{Extend the dimension of $I_\text{target}$ $T$ times}
    \State $\mathcal{I}' \gets \mathrm{LOCF}(\mathcal{I})$ \Comment{Fill empty images}
    \State $X_\tau \gets (1 - \tau)\,\mathcal{I}' + \tau\,\mathcal{I}_\text{target} + \sigma $ \Comment{Calculate the linear interpolation }
    \State
      $\mathcal{L}_{\mathrm{FM}} \gets
         \left\|\,v_\theta\bigl(\tau,\,X_\tau\bigr)\;
           -\;(\mathcal{I}_\text{target}-\mathcal{I}')
         \right\|^2$ \Comment{Calculate the velocity loss}
    \State Update $\theta \;\leftarrow\;\mathrm{AdamW} \left(  \nabla_\theta \mathcal{L}_{\mathrm{FM}} \right)$
  \EndWhile
  \State \Return $v_\theta$
    \If{inference}
\State Initialize $X_0 \gets \mathcal{I}'$
\State Define integration grid $\{\tau_0 = 0, \dots, \tau_N = 1\}$ with $n$ steps 
\State $\hat{X}_{0:N} \gets \mathrm{ODEInt}\big(v_\theta, X_0,\; \{\tau_0, \dots, \tau_N\}\big)$ \Comment{numerically integrate the network}
  \State \Return $\hat{X}_N$
\EndIf
\end{algorithmic}
    \label{alg:tfm}
\end{algorithm*}

\subsubsection{Practical Implementation}
For datasets which are regular, this grid embedding is simply achieved by randomly masking some entries.
The occupancy operator also highlights one key trade-off issue: if $\Delta$ is too large, we possibly overwrite some images. 
If $\Delta$ is too small, our context window is also very small if we keep within a budget of $K$ tensor size.

\section{Further Ablations}\label{subsec:ablations}

\subsection{Critical Ablations}

\begin{table}[!htbp]
    \centering
        \caption{\textbf{Ablation Results for discrete \CTFM{} on ACDC:} 
        This table compares \TFM{} under different design changes, showing the performance under each scenario.
        The ablations were done on an ACDC validation set.
        We evaluate the effect of using a more lightweight version of the U-Net which does not use attention('No Att').
        Instead, $\tau$ and image embeddings are merged via concatenation in the bottleneck. 
        We also compare aggregating via the mean and the last image, but these results are only for inference. Training is still done the same way.
        Third, we compare LOCF with the alternative of using the image sequences $\mathcal{I}$ as they are given, which notably reduces performance.
         *Limiting the model to only see \IB{} during training and perform FM on this is unstable, which highlights the importance of temporal context.
        }
    \label{tab:ablation_different_design_choices}
\begin{tabular}{llll}
\toprule
Change &  NRMSE ($\downarrow$) & SSIM ($\uparrow$) & PSNR ($\uparrow$)  \\
\midrule
\multirow[t]{3}{*}{Att U-Net \&  Mean }  & 0.0261 & 96.04 &  32.30  \\ 
 No Att: Mean & 0.0270 & 95.77 & 31.88 \\
 No Att: Last & 0.0271 & 95.77  &  31.87  \\
No LOCF + Masking & 0.0380 & 95.55 & 31.20\\
No LOCF & 0.0444 & 90.92 &  27.30 \\
\midrule
LCI &0.0380&93.50&29.49\\
\bottomrule
\end{tabular}
\end{table}
In Table~\ref{tab:ablation_different_design_choices} we see the different design choices of the discrete \CTFM{}. 
Notably, LOCF or masking (which is training only on the given frames), archives the highest boost versus the baseline.
We note that the masked training takes longer to converge, but yields a similar, albeit not the same performance.

\subsection{Zero-shot robustness to masking order}
\begin{figure}[!htbp]
    \centering
    \includegraphics[width=0.75\linewidth]{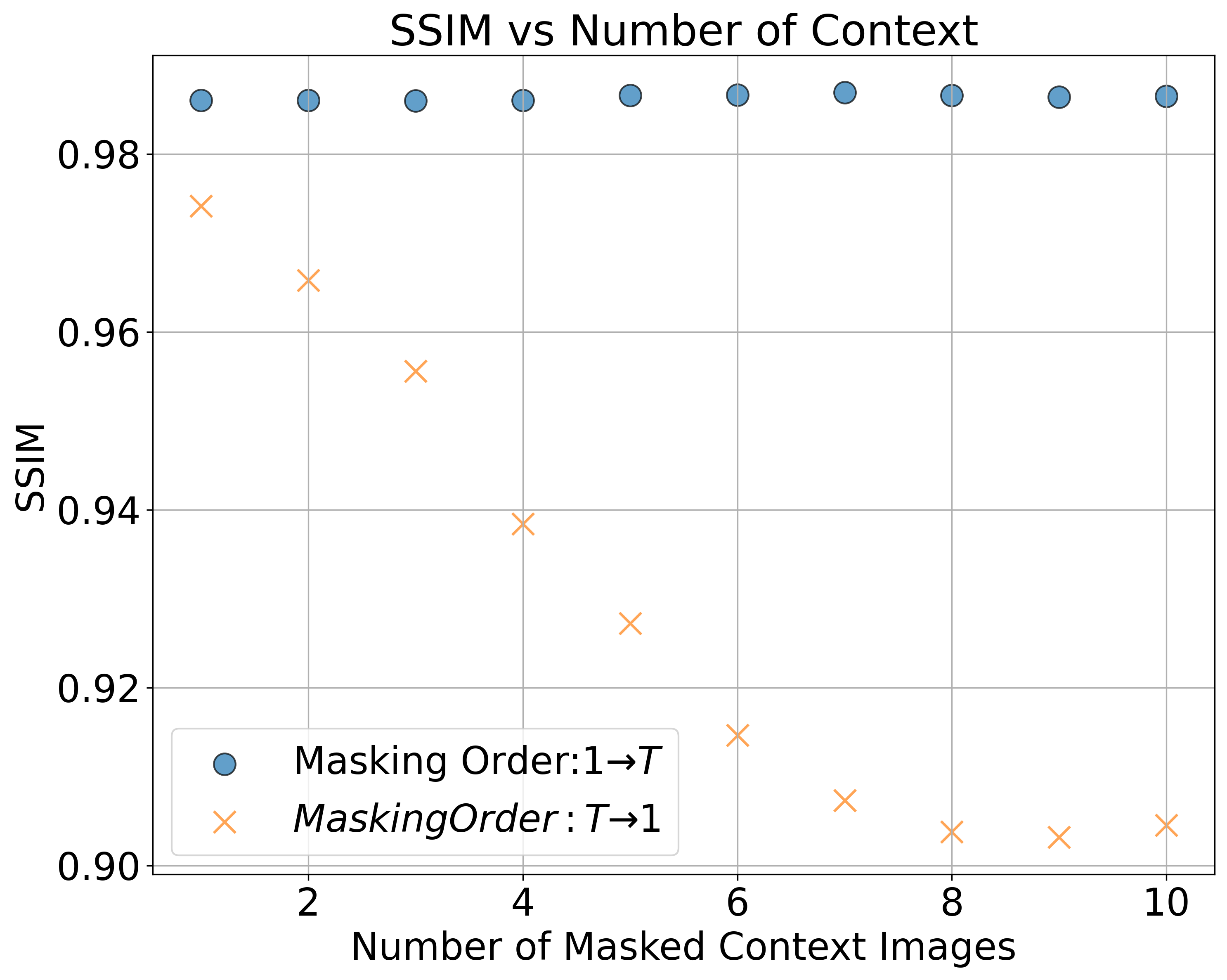}
    \caption{\textbf{Zero-shot masking of early vs. late context frames on ACDC.}
    We compare \CTFM{}'s prediction quality when masking context frames at inference time.
    In the early masking protocol ($1\to T$), we remove the earliest context frames first, in the late masking protocol ($T\to 1$), we remove the frames closest to the target. The x-axis denotes the total number of masked frames. 
    Masking frames near the target causes a clear, monotonic drop in SSIM, confirming that \CTFM{} relies on temporally proximal information.
    Masking from earlier frames leads only to a slight degradation, and shows a weak U-shaped behavior at larger  mask counts.    
    }
    \label{fig:front_back_masking}
\end{figure}

In Figure~\ref{fig:front_back_masking}, we evaluate how prediction quality changes when masking context frames from early to late.
As expected, increasing the temporal gap between the last available frame and the target leads to a monotonic decrease in SSIM.
Interestingly, when the final frame remains present and only earlier frames are removed, the curve becomes U-shaped: performance peaks when roughly five context frames are masked.
This aligns with the masking distribution used during training, where such levels of sparsity were most common.
This also suggests that the model is optimized for the typical irregular pattern seen during training, and the fully unmasked case is not the easiest \textit{in the zero-shot} evaluation.

\subsection{Hyperparameter Ablations}
We did ablation studies on one ACDC validation subset. 

\begin{table}[H]
\centering
\small
\caption[\CTFM{} Feature Size Ablation]{Ablation on feature size (FS) for the proposed method, compared to the last context image (\IB{}) baseline. Values in parentheses indicate the difference relative to \IB{}. Larger FS values generally improve NRMSE, SSIM, and PSNR, with FS = 32.0 achieving the best overall performance.}
\setlength{\tabcolsep}{6pt}
\label{tab:ablation_feature_size}
\begin{tabular}{lcccc}
\toprule
Feature Size & NRMSE ($\downarrow$) & SSIM ($\uparrow$) & PSNR ($\uparrow$) & Max Mem[$GB$] \\
\midrule
LCI & 0.038 & 93.50 & 29.49 & - \\
\midrule
FS=8.0 & 0.027 (-0.012) & 95.94 (+2.38) & 32.07 (+2.74) & 2.70 \\
FS=16.0 & 0.026 (-0.013) & 95.99 (+2.43) & 32.17 (+2.83) & 3.27 \\
FS=32.0 & 0.026 (-0.013) & \textbf{96.06 (+2.49)} & 32.24 (+2.90) & 7.70\\
FS=64.0 & \textbf{0.026 (-0.014)} & 95.97 (+2.41) & \textbf{32.35 (+3.01)} & 11.21 \\
\bottomrule
\end{tabular}
\end{table}

\begin{table}[H]
\centering
\small
\caption[\CTFM{} NFE Ablation]{
\textbf{Evaluating Metrics vs. Number of Function Evaluations:}
We evaluate how the number of function evaluations (NFEs) affects $SSIM$ performance on one ACDC validation set.
$SSIM$ increases with more evaluations and peaks at $5-10$ NFEs, after which it plateaus.
However, the improvement becomes marginal after beyond just $5$ NFEs. As trade-off we use $10$ NFE throughout.}
\setlength{\tabcolsep}{6pt}
\begin{tabular}{lccc}
\toprule
 NFE & NRMSE ($\downarrow$) & SSIM ($\uparrow$) & PSNR ($\uparrow$) \\
\midrule
LCI & 0.038 & 93.50 & 29.49 \\
\midrule
NFE=1.0 & 0.030 (-0.009) & 95.43 (+1.87) & 31.47 (+2.13) \\
NFE=5.0 & \textbf{0.025 (-0.014)} & \textbf{96.18 (+2.62)} & \textbf{32.47 (+3.13)} \\
NFE=10.0 & 0.026 (-0.013) & 96.06 (+2.49) & 32.24 (+2.90) \\
NFE=100.0 & 0.026 (-0.013) & 96.04 (+2.48) & 32.30 (+2.97) \\
NFE=200.0 & 0.025 (-0.014) & 96.18 (+2.62) & 32.47 (+3.13) \\
\bottomrule
\end{tabular}

\label{tab:ssim_vs_nfes}
\end{table}

\begin{table}[H]
\centering
\small
\setlength{\tabcolsep}{6pt}
\caption[TFM Training Noise Ablation]{\textbf{Training noise amount} validation performance ablation on training noise (TN) levels for the proposed method, compared to the last context image (LCI) baseline. Values in parentheses indicate the difference relative to LCI. Moderate TN levels (0.0--0.05) consistently improve NRMSE, SSIM, and PSNR, with TN = 0.0 yielding the best overall performance.}
\label{tab:ablation_training_noise}

\begin{tabular}{lccc}
\toprule
Training Noise & NRMSE ($\downarrow$) & SSIM ($\uparrow$) & PSNR ($\uparrow$) \\
\midrule
LCI & 0.039 & 93.53 & 29.42 \\
\midrule
TN=0.0 & 0.026 (-0.013) & 96.06 (+2.49) & 32.24 (+2.90) \\
TN=0.01 & \textbf{0.026 (-0.014)} & \textbf{96.12 (+2.56)} & 32.32 (+2.98) \\
TN=0.025 & 0.026 (-0.013) & 96.06 (+2.49) & \textbf{32.34 (+3.00)} \\
TN=0.05 & 0.026 (-0.013) & 96.04 (+2.47) & 32.26 (+2.92) \\
TN=0.1 & 0.026 (-0.013) & 96.06 (+2.49) & 32.24 (+2.90) \\
TN=0.3 & 0.026 (-0.013) & 95.78 (+2.21) & 32.26 (+2.92) \\
\bottomrule
\end{tabular}
\end{table}
In Table ~\ref{tab:ablation_training_noise}, we see the different effects of training noise on the performance. 
While we find that $0.01$ seems to yield the best performance, we kept the noise at $0$, in order to compare to other baselines, which were not trained in a noisy fashion.

\subsection{Runtime and Memory Usage} \label{subsec:runtime_memory_usage}
To provide a transparent comparison of computational efficiency, we report wall-clock training time and peak GPU memory for a single run of each baseline. 
All experiments were run on the same hardware with batch size~4. 
Table~\ref{tab:runtime_memory} shows that \CTFM{} achieves competitive runtime and memory usage compared to other spatio-temporal baselines, while diffusion-based approaches remain significantly more expensive (see main text). 

\begin{table}[h]
\centering
\begin{tabular}{lcc}
\toprule
Method & Runtime [$10^3$s] & Max Memory [GB] \\
\midrule
ViViT     & 16  & 9.0 \\
NODE+LSTM        & 60  & 6.6  \\
SimVP & 20 & 5.2 \\
ConvLSTM  & 14  & 3.5 \\
\oursrow
\CTFM{} irregular   & 18  & 7.7 \\
\oursrow
\CTFM{} continuous      & 12  & 7.0 \\
\bottomrule
\end{tabular}
\caption{Wall-clock runtime in thousands of seconds and maximum memory usage for a single run. }
\label{tab:runtime_memory}
\end{table}

\subsection{\emph{Image to image comparison}} \label{subsec:image_to_image}
\begin{figure}
    \centering
        \caption{\textbf{Image to Image - ACDC.} Efficiency–performance trade-off on ACDC: PSNR (↑) vs. training time (GPU hours, log scale). The red dashed line marks the LCI baseline; red/blue shading indicates better/worse than baseline. Bubble size encodes VRAM needs. \emph{Ours \CTFM{} (tiny)} achieves above-baseline PSNR with roughly two orders-of-magnitude less training than Latent Diffusion AE, which remains below the baseline despite substantially higher cost. } 
    \includegraphics[width=0.5\linewidth]{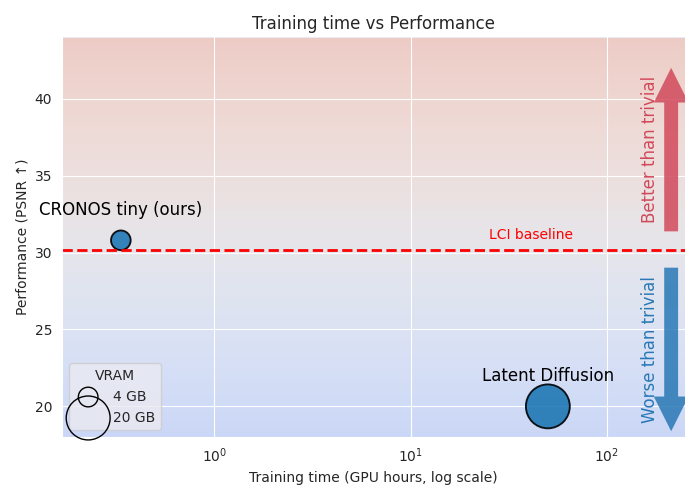}
    \label{fig:acdc_diffusion_baseline}
\end{figure}

In Figure~\ref{fig:acdc_diffusion_baseline}, we include a diffusion baseline adapted from~\cite{puglisi_brain_2025}.
Diffusion models are widely used for generative modeling, but their iterative denoising leads to orders of magnitude longer inference times and, in our setting, they do not outperform the simple \IB{} heuristic.
\textit{Most importantly}, to the best of our knowledge, \textbf{no prior} 3D continuous time diffusion approach in medical imaging \textbf{supports forecasting from multiple input volumes}. We therefore present diffusion here as a motivating image-to-image reference, but focus the remainder of our study on spatio-temporal learning (STL) methods, which are both more computationally feasible and explicitly designed for multi-context forecasting.
So, the results from ~\ref{fig:acdc_diffusion_baseline}, the inference would take roughly  35 minutes, and auto-regressive diffusion would take $5-6\,h$ in total per sampling step
\todo{update figure}

\subsection{Continuous ablation}
\subsubsection{Main Results Ablation}
For the headline comparison we further restrict to a \emph{single} context frame (image-to-image forecasting), which is markedly harder for the discrete \CTFM{} than multi-context prediction, as the last-observed carry-forward is more challenging.
We keep the same training objective and evaluation protocol; see Tab.~\ref{tab:val_loft_tfm_lci} for metrics under this stricter setting.
We observe that the continuous \CTFM{} is better than \IB{}, while the discrete one is not. 
Even stronger prediction can be seen in \ref{tab:break_continuous}.

\subsubsection{Additional Ablations for use of continuous time}
To isolate the effect of explicit time, we construct a continuous setting where only the continuous variant has access to time information.
We use a 4-frame grid, where we uniformly sample 2 images.
From the missing future frames, we predict a single future frame. 
Here, the discrete model receives \emph{no timestamps}.
The continuous model receives the two context times and the target time during inference.
All else is being equal (architecture, training regime, etc.).
Results in Tab.~\ref{tab:break_continuous} show that explicit timestamps are necessary in this protocol.

\begin{table}
\vspace{-1.0em} 
\centering
\caption{\textbf{Continuous conditioning improves sequence-to-image forecasting.}
Comparison of the LCI baseline with \CTFM{} under discrete (implicit time) and continuous (explicit timestamps) settings.
Metrics are reported as SSIM$\uparrow$, PSNR$\uparrow$, NRMSE$\downarrow$.
The continuous variant attains the best scores across all metrics, indicating that real-valued time conditioning is effectively exploited.}

\begin{tabular}{lccc}
\toprule
Method &  SSIM $\uparrow$ & PSNR $\uparrow$ & NRMSE $\downarrow$ \\
\midrule
LCI (baseline) & 0.94997 & 31.727 & 0.03038 \\
\midrule
\CTFM{} discrete              & 0.96671 & 33.534 & 0.02211 \\
\oursrow
\CTFM{} continuous            & 0.97814 & 36.007 & 0.01670 \\
\bottomrule
\end{tabular}
\label{tab:break_continuous}
\end{table}

\section{Reproducibility}\label{sec:reproducibility}
We will describe the experiments to further facilitate reproducibility.

\subsection{Experimental Details}\label{subsec:exp_details}
All methods were trained with AdamW and a cosine-annealed learning-rate schedule, using a batch size of 4.
The learning rate was fixed at $1\mathrm{e}{-4}$ for all experiments.
For \CTFM{}, we used 10 integration steps during inference, see appendix (~\ref{subsec:ablations}) for ablations.
To ensure fair comparison, we ran each experiment three times with different validation splits and the same random seed within each split.
We trained each model for the same number of epochs, selecting the checkpoint with the best NRMSE.
We report NRMSE, PSNR and SSIM for each experiment.
Further details can be found ~\ref{subsec:random_masking}, such as how we fix the random masking for ACDC and ISLES.
Furthermore, we use \ib{} (\IB{}), a heuristic that serves as strong heuristic baseline.
This is simply the last available image in the sequence.
This baseline is medically motivated, as it serves as a part of medical decision making when looking at longitudinal series (see~\cite{therasse_new_2000}).
Furthermore, in a setting with slowly changing anatomy, this is surprisingly strong.

\subsection{Random Masking}\label{subsec:random_masking}
For ACDC and ISLES, we randomly omit context images during both training and validation, to simulate irregular sampling. 
Since we believe we are the first to benchmark methods in this very specific irregular setting, we \textbf{highlight a potentially grave pitfall}:
If validation masks are resampled at each validation epoch, even with a fixed seed, the masking evaluation metrics change every time, which is exacerbated by our small validation set .
This variability affects even the trivial \IB{} baseline and makes "best" epoch selection arbitrary.
Since the validation set is small, context sequences can be extremely sparse or dense, causing the \IB{} baseline's performance to fluctuate drastically\footnote{In natural imaging, validation sets are much larger, so random fluctuations are less severe. In medical imaging, however, smaller validation sets make these fluctuations significant.}.
To avoid this issue, we generate one fixed set of masks per split (using a single seed) and reuse those exact masks for every model at every validation epoch.
This ensures consistent validation conditions, meaningful epoch selection, and fully reproducible and interpretable validation results.
In all cases, models were selected by the lowest validation $MSE$ and then evaluated on the held-out test set.

\subsection{Network architecture}\label{subsec:network_architecture}
Our default network $v_\theta$ is a four-scale 3D U\!-\!Net with residual blocks and FiLM time conditioning.
The input $X\!\in\!\mathbb{R}^{B\times (C\times T)\times D\times H\times W}$ (typically $C{=}1$) stacks $T$ context volumes along channels and is linearly mixed by a $1{\times}1{\times}1$ Conv3D to $C_{\text{stem}}{=}32$. 
The encoder uses channel expansion rates $[1,1,2,4]$ with one ResBlocks per scale (Conv3D($k{=}3$)–GN(8)–SiLU–Conv3D($k{=}3$)–GN(8), with a $1{\times}1{\times}1$ projection on the skip if needed); downsampling is Conv3D($k{=}3$, $s{=}2$).
The bottleneck has one ResBlocks and optional windowed self-attention over flattened spatial tokens (window $8^3$, $h{=}4$ heads, head dim $64$), followed by a final ResBlock. The decoder upsamples trilinearly by $2$, applies Conv3D($k{=}3$), concatenates the encoder skip, and mirrors the two-ResBlock pattern with channel width equal to the reversed down blocks..
FiLM modulation is injected after the first GN in every block: $\hat{h}=\alpha_\ell\!\odot\!\mathrm{GN}(h)+\beta_\ell+h$, where $(\alpha_\ell,\beta_\ell)$ come from an MLP over a fixed-dimensional conditioning code; either Fourier features of the scalar flow step $\tau\!\in[0,1]$ for the irregular setting, or the summed Fourier encodings of real timestamps $\sum_{i=1}^T\gamma(t_i)/T$ for the continuous setting.
All convolutions use padding to preserve spatial sizes; normalization is GroupNorm(8), activation is SiLU, and skips are concatenative with a $1{\times}1{\times}1$ channel-align conv. 
The head is Conv3D($32{\to}32$, $k{=}3$)–SiLU–Conv3D($32{\to}1$, $k{=}1$) producing a single-channel voxelwise velocity field.

\section{Addenda}
\subsection{Use of LLMs}\label{subsec:llm}
An LLM was used to help rephrase and polish this work.

\end{document}